%% file: main.tex
\PassOptionsToPackage{hyphens}{url}
\PassOptionsToPackage{type1}{XCharter}
\documentclass{template}
\usepackage[T1]{fontenc}
\ifxetex
  \microtypesetup{tracking=false}
\fi

\usepackage{newunicodechar}
\newunicodechar{Ł}{\L}
\usepackage{float}
\AddToHook{cmd/ps@firststyle/after}{%
  \fancyhead[L]{\includegraphics[height=37pt]{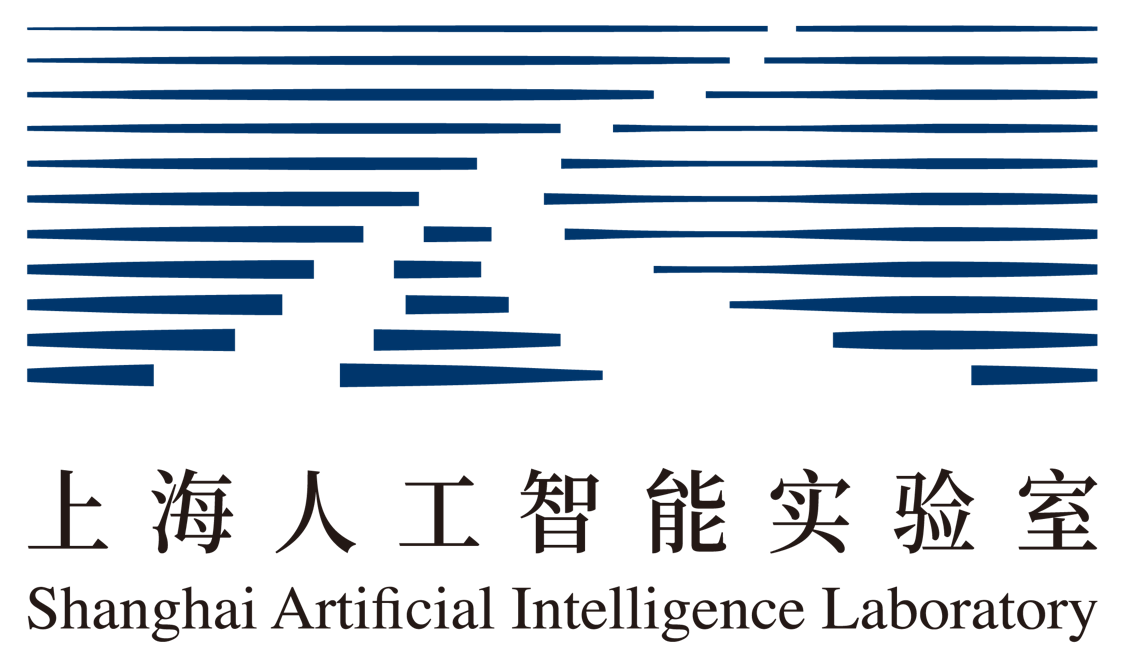}}%
}
\hypersetup{linkcolor=red,citecolor=Cerulean,filecolor=magenta,urlcolor=NavyBlue}
\renewcommand{\paragraph}[1]{\vspace{1mm}\noindent\textbf{#1}}
\newcolumntype{Y}{>{\raggedright\arraybackslash}X}
\usepackage[fixed]{fontawesome5}
\usepackage{simpleicons}
\usepackage{tikz}
\usetikzlibrary{arrows.meta}
\newcommand{\archspace}{ArchSpace}
\input{figures/icons.tex}

\newcommand\blfootnote[1]{%
  \begingroup
  \renewcommand\thefootnote{}\footnote{#1}%
  \addtocounter{footnote}{-1}%
  \endgroup
}

\title{ArchSpace: Recording, Reviewing, and Referencing LLM Architecture Research}
\author[1,4]{Jie~Wang}
\author[1]{Tong~Zhu}
\author[1,4]{Jiahao~Kuang}
\author[1]{Zhe~Xu}
\author[1,3]{Mowen~Ruan}
\author[1,3]{Yunkai~Chai}
\author[1,3,5,8]{Yuliang~Liu}
\author[1,2]{Dehong~Zheng}
\author[1]{Rubin~Wei}
\author[1,7]{Yu~Cheng}
\author[1,6]{Bowen~Zhou}
\author[1,2,$\dagger$]{\authorcr Tao~Ji}
\author[1,3,$\dagger$]{Zhouhan~Lin}
\author[1,2]{Qi~Zhang}
\affil[1]{Shanghai AI Laboratory}
\affil[2]{Fudan University}
\affil[3]{Shanghai Jiao Tong University}
\affil[4]{East China Normal University}
\affil[5]{Shanghai Innovation Institute}
\affil[6]{Tsinghua University}
\affil[7]{Nanyang Technological University}
\affil[8]{Nanjing University}
\input{sections/00_abstract.tex}

\begin{document}
\blfootnote{$\dagger$ Corresponding authors.}
\maketitle
\input{figures/literature_survey_figure.tex}
\clearpage
\input{sections/01_intro.tex}
\input{sections/02_community.tex}
\input{sections/03_framework.tex}

\input{sections/04_baseline_study.tex}
\input{sections/05_long_training.tex}
\input{sections/06_related_work.tex}
\input{sections/08_conclusion.tex}
\clearpage
\bibliography{references}
\clearpage
\appendix
\input{sections/10_appendix_configurations.tex}

\input{sections/11_appendix_baseline_recipes.tex}

\input{sections/12_appendix_scaling_laws.tex}

\input{sections/13_appendix_formula_ablations.tex}
\input{sections/15_appendix_long_training.tex}

\input{sections/16_appendix_literature_survey.tex}
\end{document}

%% file: figures/icons.tex
\newcommand{\archspaceicon}{%
  \raisebox{-.125em}{\includegraphics[trim=4bp 9bp 4bp 9bp,clip,height=1em]{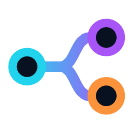}}%
}
\newcommand{\githubicon}{\faGithub}
\newcommand{\dataseticon}{\faDatabase}
\newcommand{\ideaicon}{%
  \tikz[x=1em,y=1em,baseline=-.35em,line width=.5pt,line cap=round]{%
    \node[anchor=center,inner sep=0pt] at (0,0) {\faLightbulb[regular]};
    \draw (0,.62) -- (0,.78)
      (-.36,.51) -- (-.47,.64) (.36,.51) -- (.47,.64)
      (-.55,.17) -- (-.72,.17) (.55,.17) -- (.72,.17);
  }%
}
\newcommand{\issueicon}{\raisebox{-.1em}{\includegraphics[height=1em]{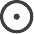}}}
\newcommand{\discussionicon}{\faComments[regular]}
\newcommand{\pricon}{\includegraphics[height=1em]{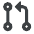}}
\newcommand{\mergedicon}{\includegraphics[height=1em]{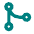}}
\newcommand{\verifiedicon}{\faCheckCircle}
\newcommand{\codeicon}{\faCode}
\newcommand{\checkpointicon}{\faSave}
\newcommand{\conclusionicon}{\faFile*}
\newcommand{\wandbicon}{\simpleicon{weightsandbiases}}
\newcommand{\huggingfaceicon}{%
  \kern.1em\raisebox{-.25em}{\includegraphics[trim=24bp 32bp 24bp 32bp,clip,height=1em]{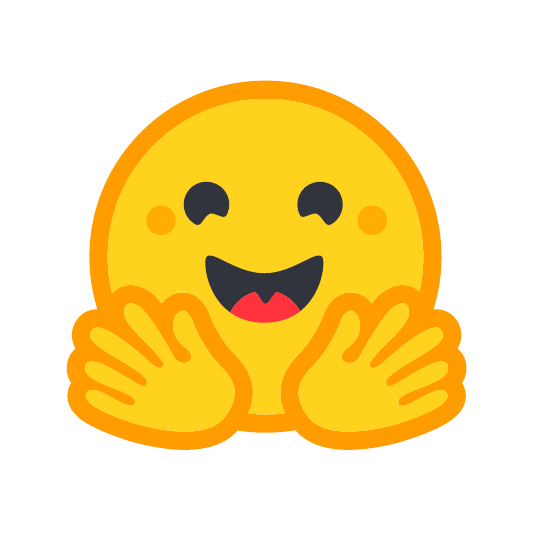}}\kern-.1em%
}

%% file: sections/00_abstract.tex
\begin{abstract}
Model architecture shapes the scaling behavior, training efficiency, and inference cost of LLMs, making it a central focus of both industry and academia. In industry, teams such as those behind DeepSeek, Qwen, and Kimi routinely introduce architectural innovations, yet public reports rarely document which ideas succeeded, which were abandoned, and why. In academia, limited compute and inconsistent baselines often restrict evaluations to small-scale, single-point comparisons, making it difficult to establish whether improvements hold at larger scales or translate into broader impact.
\archspace{} addresses both gaps as a public platform for \emph{Recording, Reviewing, and Referencing Architecture Research}: each proposal is recorded with its experiments and reasoning, reviewed by the community, and referenced by later work. To evaluate ideas before costly large-scale training, \archspace leverages scaling laws to predict model loss at larger scales from smaller experiments, a practice widely adopted in industry. We provide initial references that make the platform usable: Qwen3.5 and OLMo3 baselines, fitted scaling laws, and a reusable training recipe. Ideas validated through this review can in turn become new baselines, and \archspace{} thereby aims to broaden the range of ideas that shape the next generation of models.
\par\medskip
{\centering\normalfont
\archspaceicon\enspace\href{https://www.archspace.live/}{https://www.archspace.live/}\quad
\href{https://github.com/InternLM/archspace}{\githubicon\enspace GitHub}\quad
\href{https://huggingface.co/datasets/ArchSpace-Collection/350B-Pipeline-Dataset}{\dataseticon\enspace Training dataset}\\
\href{https://huggingface.co/collections/ArchSpace-Collection/qwen35-baseline}{\huggingfaceicon\enspace Qwen3.5 baseline collection}\quad
\href{https://huggingface.co/collections/ArchSpace-Collection/olmo3-baseline}{\huggingfaceicon\enspace OLMo3 baseline collection}\par}
\end{abstract}

%% file: figures/literature_survey_figure.tex
\begin{figure}[H]
\centering
\includegraphics[width=\linewidth]{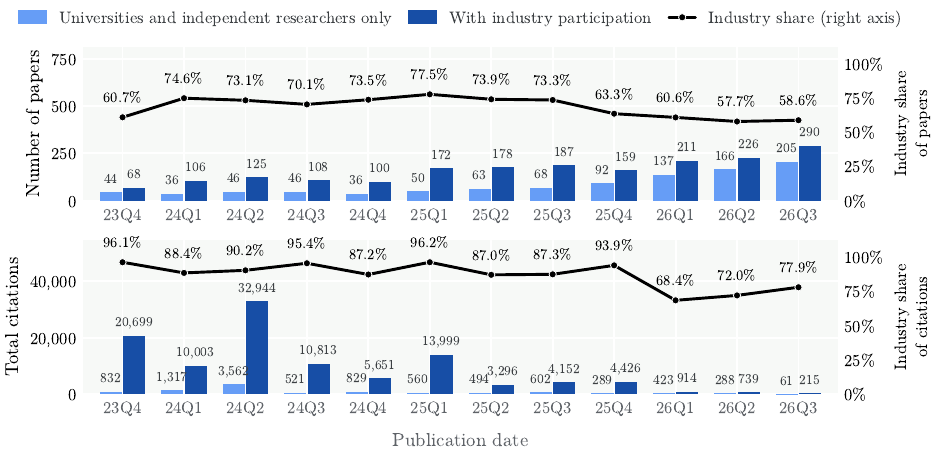}
\caption{Academic architecture research receives a disproportionately small citation share, motivating \archspace{}'s goal of broader visibility and recognition.}
\label{fig:literature_survey}
\end{figure}

%% file: sections/01_intro.tex
\section{Introduction}
\label{sec:intro}
% Role: opening; architectural innovations come from experiments and design choices whose public records would help subsequent research build on successful and unsuccessful attempts.
Architecture forms the foundation of a large language model (LLM), shaping its performance, scaling behavior, and efficiency during training and inference. When teams such as DeepSeek, Qwen, and Kimi release new open models, they also introduce architectural changes and innovations. Validating these designs involves costly training experiments. However, the experiments and trade-offs behind most models remain inaccessible, leaving the community with little account of which ideas succeeded, which were abandoned, and why. An open and complete record of validation would therefore be as valuable to subsequent research as a new idea itself.

% Role: visibility problem; connect the value of research records to discovery and further use of the documented ideas, then show the citation imbalance.
The value of such records also depends on whether researchers discover the documented ideas and build on them. However, much of the attention to new architectures is concentrated on research from industry: our survey\footnote{See Appendix~\ref{app:survey_methodology} for survey details.} shows that academic work accounts for about a third of the papers but receives less than 10\% of the citations (Figure~\ref{fig:literature_survey}). Academic ideas need greater visibility and proper validation so that the community can discuss these ideas and draw inspiration from them, and even explore new architectural designs on these bases.

% Role: contribution; ArchSpace ties the public record to a validation framework so recorded gains reflect the architecture itself.
We therefore introduce \archspace{}, a public platform for \emph{Recording, Reviewing, and Referencing Architecture Research}. Each proposal is recorded with its experiments and reasoning, reviewed by the community to check whether the results support the conclusion, and referenced by later work. To evaluate the architectural changes described in each proposal, the validation framework uses scaling laws~\citep{kaplan2020,hoffmann2022} to compare models across multiple compute budgets, a systematic and standard practice widely used in industry~\citep{gpt4,llama3,deepseekllm2024,minimax01,hu2024minicpm,mistrallarge4}.

% Role: evidence and purpose; reusable baselines and the community vision close on the opening problems.
As initial references, we provide Qwen3.5 and OLMo3 baselines. Validated architecture innovations can in turn become new baselines for later work. \archspace{} aims to build a community around these contributions, where the research process is openly documented and a wider range of ideas can gain visibility and recognition.

%% file: sections/02_community.tex
\section{An Open Community for Architecture Research}
\label{sec:public_review}
% Role: connect community participation to recording, reviewing, and referencing research.
\archspace{} aims to build an open community for \emph{Recording, Reviewing, and Referencing Architecture Research}. Each proposal is recorded with its experiments and reasoning, reviewed by the community to check whether the results support the conclusion, and referenced by later work.

% Role: contribution workflow, supporting record, public review, and reuse.
Contributing to \archspace{} begins with an Architecture Proposal issue, which states the proposed change, its rationale, the baseline, and the validation plan (Figure~\ref{fig:pipeline}). Community discussion helps contributors refine the architectural idea and plan a fair comparison with the baseline. Contributors then implement and evaluate the proposed architecture under the validation framework (Section~\ref{sec:validation_framework}) and submit their work for public review through a pull request (PR) linked to the issue. The PR links to the implementation, configurations, checkpoints, training logs, evaluation results, and reproduction instructions. It also records failed trials and search choices so readers can trace how the conclusion was reached. Community reviewers then use this record to check whether the implementation matches the proposal, the comparison is reproducible, and the experiments support the conclusion. The verified PR is then merged into \archspace{} for future reference, and the proposed architecture can serve as a new baseline.

\input{figures/pipeline.tex}

% Role: the vision for PRs as scholarly contributions and a shared foundation for subsequent research.
Our broader aim is for reviewed PRs to have the same standing as traditional research papers and become the primary way to communicate and cite contributions in \archspace{}. A PR combines a complete account of the contribution with direct links to the resources needed to reproduce and extend the work. Follow-up PRs can link to earlier contributions, so researchers can trace how architectural ideas develop across studies. By bringing these contributions together, \archspace{} aims to provide a shared foundation for subsequent architecture research.

%% file: figures/pipeline.tex
% Story: open a proposal -> Phase I (one of three settings) -> Phase II (two stages) -> contribute.
\begin{figure}[p]
\centering
% PAIR explorable palette, shared with scripts/plot_style.py; budget shades run light (1e18) to dark (1e20).
\definecolor{budgetA}{HTML}{A2DA37}
\definecolor{budgetB}{HTML}{32B67A}
\definecolor{budgetC}{HTML}{24868E}
\definecolor{budgetD}{HTML}{3A548C}
\definecolor{budgetE}{HTML}{481769}
\definecolor{asgray}{HTML}{5F6368}
\definecolor{flowgray}{HTML}{80868B}
\definecolor{pairtext}{HTML}{3C4043}
\definecolor{pairpanel}{HTML}{F7F9F7}
\definecolor{pairtint}{HTML}{EDF4EC}
\definecolor{pairedge}{HTML}{C6D7CE}
\definecolor{pairteal}{HTML}{008080}
\definecolor{pairpurple}{HTML}{6E14C2}
\definecolor{pairpurplepale}{HTML}{F6ECFF}
\definecolor{pairpurplelight}{HTML}{EDDAFF}
\small\color{pairtext}
\begin{tikzpicture}[x=1cm,y=1cm,>=Latex,
  flow/.style={->,line width=1.1pt,draw=flowgray},
  card/.style={rounded corners=4pt,line width=.6pt},
  tag/.style={font=\footnotesize\bfseries,anchor=base west,inner sep=0pt},
  stage title/.style={font=\large\bfseries,anchor=base west,inner sep=0pt},
  line/.style={anchor=base west,inner sep=0pt},
  note/.style={font=\footnotesize\itshape,text=asgray,inner sep=1pt}]
\useasboundingbox (0,.05) rectangle (16,-19.8);

% Role: develop an idea through a proposal issue and discussion.
\filldraw[card,fill=pairpanel,draw=pairedge] (0,0) rectangle (16,-1.1);
\node[tag,text=asgray] at (.35,-.47) {START};
\node[line,font=\footnotesize\bfseries] (start-idea) at (1.9,-.47) {\ideaicon\enspace Design idea};
\node[anchor=base east,inner sep=0pt,font=\footnotesize\bfseries] (start-discussion) at (15.65,-.47)
  {\discussionicon\enspace Community discussion};
% Center the middle label between the outer labels to fill the row with equal arrows.
\path (start-idea.base east) --
  node[midway,anchor=base,inner sep=0pt,font=\footnotesize\bfseries] (start-issue)
    {\issueicon\enspace Architecture Proposal issue}
  (start-discussion.base west);
\coordinate (start-flow) at (0,-.35);
\foreach \source/\target in {start-idea/start-issue,start-issue/start-discussion} {
  \draw[flow,line width=.8pt,shorten <=3pt,shorten >=3pt]
    (\source.east |- start-flow) -- (\target.west |- start-flow);
}
\node[line,font=\footnotesize] at (.35,-.9)
  {State the change, baseline, and validation plan; assess novelty, feasibility, and experimental design.};
\draw[flow] (8,-1.1) -- (8,-1.45);

% Role: Phase I, five compute budgets and three alternative settings.
\filldraw[card,fill=pairpanel,draw=pairteal!45] (0,-1.45) rectangle (16,-11.45);
\node[tag,text=pairteal] at (.35,-1.92) {PHASE I};
\node[stage title] at (2.25,-1.92) {Scaling across compute budgets};
\node[line] at (.35,-2.45)
  {Same data and tokenizer as the baseline.};
\node[line,font=\footnotesize\bfseries] at (.35,-2.98) {Scaling ladder (training FLOPs)};
\foreach \i/\budget/\col in {0/{$1\times10^{18}$}/budgetA,1/{$3\times10^{18}$}/budgetB,
    2/{$1\times10^{19}$}/budgetC,3/{$3\times10^{19}$}/budgetD,4/{$1\times10^{20}$}/budgetE} {
  \pgfmathsetmacro{\xx}{.35+3.1*\i}
  \filldraw[rounded corners=3pt,fill=white,draw=pairedge] (\xx,-3.15) rectangle +(2.9,-.5);
  \fill[\col] (\xx+.55,-3.4) circle[radius=.13cm];
  \node[anchor=west,inner sep=0pt] at (\xx+.9,-3.4) {\budget};
}
% A single fork: the three settings are alternatives.
\draw[line width=.8pt,draw=flowgray] (8,-3.65) -- (8,-4.05);
\draw[line width=.8pt,draw=flowgray] (2.833,-4.05) -- (13.167,-4.05);
\foreach \xx in {2.833,8,13.167} {\draw[flow,line width=.8pt] (\xx,-4.05) -- (\xx,-4.35);}
\node[note,anchor=west] at (8.1,-3.85) {choose one setting};

% Setting cards with their fitted scaling law and approximate training cost.
\foreach \xx/\num/\name/\sizes/\hp/\cost/\law in {
    .35/1/Baseline-optimal points/baseline-optimal/reused/{$1.4\times10^{20}$}/{$\widehat L(C)$},
    5.517/2/IsoFLOP size sweep/4--6 per budget/reused/{$7.2\times10^{20}$}/{$\widehat L(M,D)$},
    10.683/3/{LR/BS-tuned IsoFLOP}/4--6 per budget/{$4\times4$ grid}/{$1.2\times10^{22}$}/{$\widehat L(M,D,\eta,B)$}} {
  \filldraw[card,fill=white,draw=pairedge] (\xx,-4.35) rectangle +(4.967,-6.1);
  \fill[rounded corners=4pt,fill=pairtint] (\xx,-4.35) rectangle +(4.967,-.8);
  \fill[fill=pairtint] (\xx,-4.75) rectangle +(4.967,-.4);
  \node[line,font=\footnotesize\bfseries,text=asgray] at (\xx+.2,-4.65) {Setting \num};
  \node[line,font=\small\bfseries] at (\xx+.2,-5.0) {\name};
  % Key--value rows span the card width: keys left, values right.
  \foreach \yy/\key/\val in {-8.55/Sizes/\sizes,-9.05/{LR, BS}/\hp,-9.55/Fit/\law,-10.05/FLOPs/\cost} {
    \node[line,font=\normalsize\bfseries] at (\xx+.2,\yy) {\key};
    \node[anchor=base east,inner sep=0pt,font=\normalsize] at (\xx+4.767,\yy) {\val};
  }
}
% Sketches are drawn at printed size, so their 7 pt labels are not rescaled.
\node[anchor=north,inner sep=0pt] at (2.833,-5.25) {\includegraphics{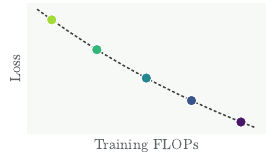}};
\node[anchor=north,inner sep=0pt] at (8,-5.25) {\includegraphics{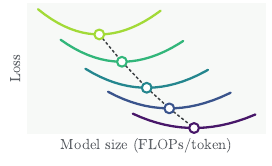}};
\node[anchor=north,inner sep=0pt] at (13.167,-5.25) {\includegraphics{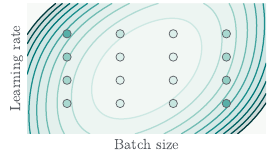}};
\node[line] at (.35,-11.05)
  {\textbf{Compare} compute--loss curves with the baseline at matched compute.};
\draw[flow] (8,-11.45) -- (8,-11.9);
\node[anchor=west,font=\footnotesize,text=flowgray] at (8.15,-11.675) {Scaling gain supported};

% Role: Phase II, two training stages, each followed by evaluation and comparison with the baseline.
\filldraw[card,fill=pairpurplepale!55,draw=pairpurple!35] (0,-11.9) rectangle (16,-16.95);
\node[tag,text=pairpurple] at (.35,-12.37) {PHASE II};
\node[stage title] at (2.25,-12.37) {Training 1B models on 150B+10B tokens};
\node[anchor=base east,inner sep=0pt,font=\small] at (15.65,-12.9)
  {\textbf{FLOPs}\enspace $1.2$--$1.5\times10^{21}$};
\node[line] at (.35,-12.9) {Same data, context schedule, and token counts as the baseline.};
% Stages: tokens, data, and context.
\foreach \xx/\stage/\data/\ctx/\evals in {.35/{150B-token pretraining}/{Dolma~3 Pool 150B sample}/{8K context}/{Base Main evaluation},
    8.3/{10B-token long-context mid-training}/{Dolma~3 Longmino Mix (50B)}/{64K context}/{Base Main and Base Long-Context evaluation}} {
  \fill[rounded corners=3pt,pairpurplelight!80] (\xx,-13.25) rectangle +(7.35,-1.55);
  \node[align=center] at (\xx+3.675,-14.025) {\textbf{\stage}\\{\footnotesize \data}\\{\footnotesize \ctx}};
  \draw[flow,draw=pairpurple,line width=.8pt] (\xx+3.675,-14.8) -- (\xx+3.675,-15.2);
  \filldraw[rounded corners=3pt,fill=white,draw=pairpurple!45,line width=.5pt] (\xx,-15.2) rectangle +(7.35,-.75);
  \node[font=\footnotesize] at (\xx+3.675,-15.575) {\evals};
}
\draw[flow,draw=pairpurple] (7.73,-14.025) -- (8.27,-14.025);
\node[line] at (.35,-16.55)
  {\textbf{Compare} Base Main results after each stage and Base Long-Context results after mid-training.};
\draw[flow] (8,-16.95) -- (8,-17.4);
\node[anchor=west,font=\footnotesize,text=flowgray] at (8.15,-17.175) {Gains confirmed};

% Role: review and merge.
\filldraw[card,fill=pairpanel,draw=pairedge] (0,-17.4) rectangle (16,-19.65);
\node[tag,text=asgray] at (.35,-17.82) {CONTRIBUTE};
\node[stage title] at (3.1,-17.82) {Records for future reference};
\foreach \xx/\artifacticon/\artifactlabel in {
    1.35/\pricon/{Pull request},
    3.4/\codeicon/{Code \& config},
    5.45/\checkpointicon/{Checkpoint},
    7.5/\wandbicon/{Training logs},
    9.55/\conclusionicon/{Conclusion}} {
  \node[anchor=center,font=\large,inner sep=0pt] at (\xx,-18.35) {\artifacticon};
  \node[anchor=base,font=\footnotesize,inner sep=0pt] at (\xx,-18.9) {\artifactlabel};
}
\draw[flow] (10.55,-18.35) -- (11.65,-18.35);
\node[anchor=south,font=\footnotesize,text=asgray,inner sep=0pt] at (11.1,-18.18) {Review};
\foreach \xx/\statusicon/\statuslabel in {
    12.65/\verifiedicon/{Verified},14.75/\mergedicon/{Merged}} {
  \node[anchor=center,font=\large,text=pairteal,inner sep=0pt] at (\xx,-18.35) {\statusicon};
  \node[anchor=base,font=\footnotesize\bfseries,text=pairteal,inner sep=0pt] at (\xx,-18.9) {\statuslabel};
}
\node[line,font=\footnotesize,text=asgray] at (.35,-19.3)
  {Future proposals can cite the evidence, reproduce comparisons, and build on the implementation.};

\end{tikzpicture}
\caption{The ArchSpace contribution workflow. Phase~I compares the proposed architecture with its baseline across the scaling ladder; the three settings differ in their reuse of model size, token count, learning rate (LR), and batch size (BS). Phase~II tests whether gains persist through longer training and evaluates downstream capabilities. Reviewed records support future reference. Here, $\widehat L$ denotes predicted loss; $C$ is training FLOPs, $M$ training FLOPs per token, $D$ the training token count, $\eta$ LR, and $B$ BS.}
\label{fig:pipeline}
\end{figure}

%% file: sections/03_framework.tex
\section{The ArchSpace Validation Framework}
\label{sec:validation_framework}
% Role: why matched comparison conditions are needed and what the framework defines.
\archspace{} standardizes architecture comparisons through a two-phase validation framework (Figure~\ref{fig:pipeline}). Phase~I tests whether the proposed architecture improves the compute--loss trend across the five budgets of the scaling ladder. Phase~II trains 1B models on 150B+10B tokens to test whether the gain persists through longer training and to evaluate downstream capabilities directly.

% Role: shared comparison conditions, training data, and their use in each phase.
To isolate the effect of the architectural change, the proposed architecture and its baseline use the same training data, tokenizer, and validation data in both phases. The \href{https://huggingface.co/datasets/ArchSpace-Collection/350B-Pipeline-Dataset}{ArchSpace training dataset} provides official OLMo~3 mixtures pre-tokenized with the OLMo~3 tokenizer~\citep{olmo3}.\footnote{Tokenizer identifier: \nolinkurl{allenai/dolma2-tokenizer}.} In our training dataset, the data for pretraining in both phases come from the \href{https://huggingface.co/datasets/allenai/dolma3_mix-150B-1025}{Dolma~3 Pool 150B sample}, and the data for Phase~II long-context mid-training come from the \href{https://huggingface.co/datasets/allenai/dolma3_longmino_mix-50B-1025}{Dolma~3 Longmino Mix (50B)}.

\subsection{Phase I: Scaling across compute budgets}
\label{sec:phase1}
% Role: why gains are compared across budgets, how scaling laws support the comparison, and the loss used for fitting.
Architectural gains can vary with the compute budget, so Phase~I validates the proposed architecture against its baseline across multiple budgets using fitted scaling laws that estimate each architecture's lowest achievable loss~\citep{kaplan2020,hoffmann2022}. The proposed architecture is considered an improvement if its compute-optimal loss curve lies on the Pareto frontier. The validation loss $L$ used for fitting is computed as the equal-weight average over the eleven domains of the original OLMo~3 validation set~\citep{olmo3} (Appendix~\ref{app:fitting}). Throughout the report, loss refers to this validation loss.

% Role: compute accounting, the model-size and training-data trade-off, and the scaling ladder used to fit each architecture's scaling law.
Let $M$ denote a model's scale in training FLOPs per token at its training context length,\footnote{Computing $M$ across architectures requires a consistent convention for counting forward and backward FLOPs. We use \href{https://github.com/NVIDIA/Megatron-LM/blob/b92c3ed70029005fd5769d2cf4c65bc3d061a74b/megatron/training/training.py\#L821-L1407}{Megatron-LM} as a reference.} $D$ the number of training tokens, and $C=MD$ the total training FLOPs. At a fixed budget $C$, increasing $M$ leaves fewer tokens for training, with $D=C/M$. To fit a scaling law for each architecture, \archspace{} sweeps the five compute budgets $C_j$ of its \emph{scaling ladder}: $1\times10^{18}$, $3\times10^{18}$, $1\times10^{19}$, $3\times10^{19}$, and $1\times10^{20}$ FLOPs. The largest-budget runs dominate the training cost, while runs at smaller budgets add modest compute. Training models across the scaling ladder therefore enables systematic comparisons between architectures at little additional computational cost.

% Role: differences in search scope and total cost, why baseline searches can be reused, and how to choose a setting.
We provide three settings that differ in the scope of hyperparameter search. Broader searches require more training runs and therefore a larger total FLOPs budget. When the proposed architecture is similar to its baseline, the optimal learning rate (LR) and batch size (BS) can also be similar for the two architectures. The baseline's search results can then be reused to reduce the cost of validation. The choice of setting therefore depends on the extent of the architectural change and the available compute.

% Role: an overview of the three settings before their detailed descriptions.
At each budget of the scaling ladder, Setting~1 reuses the baseline's compute-optimal settings, including model size, training token count, LR, and BS. Setting~2 performs an IsoFLOP sweep over model size and training token count while reusing the baseline's LR and BS. Setting~3 additionally tunes LR and BS during the IsoFLOP sweep (Figure~\ref{fig:pipeline}).

\paragraph{Setting 1: baseline-optimal settings.}
% Role: selecting the model and training settings, then fitting loss as a function of compute.
Setting~1 trains one model per budget. The baseline's scaling law gives the target model scale $M^*(C_j)$ (Section~\ref{sec:analysis}). A realizable model of the proposed architecture rarely has exactly this scale, so the model with training FLOPs per token closest to the target should be selected. The selected model is trained either for the token count given by the baseline's compute-optimal curve, allowing the budget to shift slightly, or at the fixed budget $C_j$, accepting a small departure from the optimal model size and training token count (Appendix~\ref{app:allocation}). LR and BS should be taken from the best measured baseline recipe at the same budget and closest model scale (Appendix~\ref{app:recipes}). The five measured losses are then used to fit a power law in compute, following \citet{kaplan2020}, with the irreducible loss term used in the Chinchilla law~\citep{hoffmann2022}:
\begin{equation}
 \widehat L(C)=L_\infty+\alpha_C C^{-\gamma_C}.
 \label{eq:compute_law}
\end{equation}
The fitting parameters are the irreducible loss $L_\infty$, amplitude $\alpha_C$, and exponent $\gamma_C$.

\paragraph{Setting 2: IsoFLOP sweep over model size and training data.}
% Role: define the IsoFLOP sweep, explain its purpose, and fit loss across model scales and token counts in Setting 2.
An IsoFLOP sweep compares models of different sizes trained with the same total compute budget, adjusting the training token count for each size~\citep{hoffmann2022}. This comparison identifies the combination of model size and training tokens that gives the lowest loss at that budget. Setting~2 searches four to six model sizes at each budget of the scaling ladder. The selected sizes should bracket the expected optimum for the proposed architecture. LR and BS should be reused from the best measured recipe of the nearest baseline model--budget group, as in Setting~1. The losses measured across these IsoFLOP sweeps are used to fit $\widehat L(M,D)$ using the coupled form in Equation~\ref{eq:coupled}. This law describes how loss varies with model scale and training token count.

\paragraph{Setting 3: LR- and BS-tuned IsoFLOP.}
% Role: initializing the LR--BS search and fitting loss across model sizes and recipes.
At each model--budget configuration in the size sweep, the LR--BS search starts from the baseline's predicted optima (Appendix~\ref{app:predicted_optima}) or, at an imposed BS, from the optimal LR at that BS (Equation~\ref{eq:conditional_lr}). The grid can be adjusted to the proposed architecture and the available compute; we use a $4\times4$ grid as an example. The losses from the sweeps over model size, LR, and BS are used to fit the full scaling law $\widehat L(M,D,\eta,B)$ in Equation~\ref{eq:joint}, where $\eta$ and $B$ denote LR and BS.

\subsection{Phase II: Training 1B models on 150B+10B tokens}
\label{sec:phase2}
% Role: connect downstream validation to the chosen model size and training design.
Phase~II tests whether gains measured on the scaling ladder persist through longer training and context extension, and evaluates downstream capabilities. The framework adopts 1B parameters as a reasonable model size for this evaluation, with pretraining on about 150B tokens at 8K context followed by long-context mid-training on 10B tokens at 64K.\footnote{Context length and BS use binary multiples ($1\mathrm{K}=2^{10}$); token and parameter counts use decimal multiples ($1\mathrm{B}=10^{9}$).}

% Role: connect the searches for pretraining and long-context mid-training to reuse of the resulting baseline recipe.
For pretraining, the LR and BS predicted by the Phase~I scaling laws provide a good starting point for hyperparameter search. In practice, the predicted optima differ from the best measured settings, so we conduct a further search over LR and BS. We then use the selected pretrained checkpoint for long-context mid-training and tune LR and BS with the pretraining settings as a reference (Section~\ref{sec:phase2_baselines}). As in Phase~I, researchers can directly reuse the tuned settings, or run their own search when additional compute is available.

% Role: connect final-checkpoint comparison to general capability evaluation, long-context evaluation, and checks for retained capabilities.
Runs are compared at the final checkpoint of each training stage, after its LR schedule completes, because intermediate rankings can reverse during LR decay (Appendix~\ref{app:rank_reversal}). Downstream capabilities are evaluated with OLMES~\citep{olmes}\footnote{\url{https://github.com/allenai/olmes}} using the evaluation suites introduced in OLMo~3~\citep{olmo3}. After pretraining, the Base Main suite tests general capabilities, involving mathematical reasoning and question answering. After long-context mid-training, evaluation focuses on the Base Long-Context suite, with RULER~\citep{ruler} as the primary development signal and HELMET~\citep{helmet} for complementary evaluation. The Base Main suite is also rerun to check whether the capabilities developed during pretraining are retained. The proposed architecture should use the same evaluation configuration as the baseline, so differences in their scores reflect only architectural differences.

%% file: sections/04_baseline_study.tex
\section{Phase I: Baseline Ladders and Scaling Laws}
\label{sec:ladders}
% Role: motivate the baseline choice and explain the two hybrid architectures.
Hybrid architectures have become a mainstream design choice for LLMs. We therefore select Qwen3.5 and OLMo3 as baselines. Qwen3.5 interleaves three Gated DeltaNet layers~\citep{yang2025gdn} with one full-attention layer~\citep{qwen35}, whereas OLMo3 interleaves three sliding-window attention layers with one full-attention layer~\citep{olmo3}.

% Role: introduce the baseline ladders and explain how they support validation.
For each architecture, we provide a \emph{baseline ladder}: a series of models trained at the five compute budgets of the scaling ladder, with LR and BS swept within each model--budget group. We fit a scaling law to each ladder's measured losses to obtain the compute-optimal model size and training token count, together with predicted optima. We then test the loss predictions on extrapolation runs at larger budgets to validate the effectiveness of the fitted scaling laws.

\subsection{Experimental setup}
\label{sec:setup}
% Role: describe the shared training recipe.
The models in both ladders are pretrained on the Dolma~3 Pool 150B sample using the shared OLMo~3 tokenizer (Section~\ref{sec:validation_framework}) and 8K-token sequences. Training combines Muon~\citep{jordan2024muon,liu2025muon} with AdamW (Appendix~\ref{app:optimizer}) under a warmup--stable--decay (WSD) schedule~\citep{hu2024minicpm}, with a square-root cooldown over the final 20\% of updates~\citep{hagele2024}.

% Role: connect the outer IsoFLOP grid and inner hyperparameter search to Setting 3 and the results.
Following Phase~I Setting~3 (Section~\ref{sec:phase1}), we conduct IsoFLOP experiments by sweeping an outer grid of model sizes and compute budgets (Figure~\ref{fig:isoflops}). Within each model--budget group, we further sweep an inner grid of LR and BS to find the best hyperparameters (Figures~\ref{fig:response_maps} and~\ref{fig:response_maps_olmo}). Appendices~\ref{app:architectures} and~\ref{app:recipes} specify the model sizes and LR--BS grids for each ladder.

\subsection{Analysis method}
\label{sec:analysis}
\paragraph{Fitting the scaling law in two stages.}
% Role: introduce the loss decomposition and its two fitting stages.
We fit a scaling law in two stages to describe how loss varies with model scale $M$ (training FLOPs per token), training token count $D$, LR $\eta$, and BS $B$:
\begin{equation}
 \widehat L(M,D,\eta,B)=L^*(M,D)+\Delta L(M,D,\eta,B),\qquad \Delta L\geq0.
 \label{eq:joint}
\end{equation}
Here the base surface $L^*(M,D)$ gives the minimum loss achievable under the given model scale and training token count. The LR--BS term $\Delta L$ gives the additional loss when LR or BS departs from its optimal value. At the predicted optimal LR and BS, $\Delta L=0$ and $\widehat L=L^*$.

% Role: explain the two fitting stages and why the base surface remains fixed.
In the first stage, we fit $L^*(M,D)$ to the lowest measured loss from each LR--BS inner grid, following OpenEuroLLM~\citep{openeuro2026}. In the second stage, we fit $\Delta L$ using the losses of all swept runs across both outer and inner grids. We hold the fitted $L^*$ fixed during this stage to preserve the IsoFLOP curves fitted to the group minima (Appendix~\ref{app:fit_order}).

% Role: define the shared fitting objective and point to its exact form.
Both stages minimize the sum of Huber losses on the differences between predicted and measured log loss. The Huber loss grows quadratically for small differences and linearly once their magnitude exceeds a threshold. Appendix~\ref{app:two_stage} gives the exact definition, threshold, and fitting objectives.

\paragraph{Modeling the minimum loss across model and data scales.}
% Role: extend the Chinchilla form with the coupling exponent used in prior work.
For the base surface, we extend the Chinchilla law~\citep{hoffmann2022}\footnote{Chinchilla uses parameter count $N$ as model scale, whereas DeepSeek LLM uses training FLOPs per token $M$~\citep{deepseekllm2024}. $N$ and $M$ are approximately interchangeable when fitting scaling laws because $M\approx6N$. We use $M$ to represent model scale because it reflects training compute more accurately.} with a coupling exponent $\gamma_{MD}$, following Skaling~\citep{skaling2026} and OpenEuroLLM~\citep{openeuro2026}:
\begin{equation}
 L^*(M,D)=L_\infty+
 \left(\alpha_M M^{-\gamma_M}+\alpha_DD^{-\gamma_D}\right)^{\gamma_{MD}},
 \label{eq:coupled}
\end{equation}
Here $L_\infty$ is the irreducible loss. The amplitudes $\alpha_M$ and $\alpha_D$ set the magnitudes of the model and data terms. The exponents $\gamma_M$ and $\gamma_D$ determine how these terms decrease as $M$ and $D$ increase, while $\gamma_{MD}$ controls their interaction. The amplitudes and exponents are all positive.

When $\gamma_{MD}=1$, the two power-law terms add directly, recovering the Chinchilla form with $M$ as model scale. We call this the \emph{additive} form and fit it as an ablation to the standard \emph{coupled} form (Appendix~\ref{app:additive}). Table~\ref{tab:coupled_parameters} gives the fitted coefficients of the coupled base surface for each baseline ladder.

\input{tables/coupled_parameters.tex}

\paragraph{Modeling loss away from the optimal LR and BS.}
% Role: give the power laws and identify their fitting parameters.
We first model the optimal LR $\eta^*$ and BS $B^*$ as power laws in $M$ and $D$, a form used in StepLaw~\citep{steplaw2025}, OpenEuroLLM~\citep{openeuro2026}, and Microsoft Law~\citep{bjorck2025}:
\begin{equation}
 \eta^*=\alpha_\eta M^{\gamma_{\eta,M}}D^{\gamma_{\eta,D}},\qquad
 B^*=\alpha_B M^{\gamma_{B,M}}D^{\gamma_{B,D}},\qquad
 h=\alpha_h M^{\gamma_{h,M}}D^{\gamma_{h,D}}.
 \label{eq:power_laws}
\end{equation}
We also model the curvature scale $h>0$ in the same way, which will be used in the LR--BS term $\Delta L$ (Equation~\ref{eq:delta_l}). The fitting parameters are $\alpha_{\square}>0$ and $\gamma_{\square,\triangle}$, with $\square\in\{\eta,B,h\}$ and $\triangle\in\{M,D\}$.

% Role: motivate cosh and the fixed LR--BS exponent before introducing the complete term.
To guide the form of $\Delta L$, we use the optimization proxy for Muon-type methods~\citep{shulgin2026,su2026scaling}. Starting from this proxy, we derive two hyperbolic cosine ($\cosh$) terms that describe its joint dependence on LR and BS in log space (Appendix~\ref{app:lrbs_term}). We therefore model the loss increase away from $\eta^*$ and $B^*$ with two cosh terms. With $\widetilde{\eta}=\ln(\eta/\eta^*)$ and $\widetilde{B}=\ln(B/B^*)$ denoting the deviations of LR and BS from their optima in log space, the LR--BS term is
\begin{equation}
 \Delta L(M,D,\eta,B)=2h\left[\cosh(\widetilde{\eta}-\gamma_{\eta,B}\widetilde{B})-1\right] +2h\rho\left(\cosh\widetilde{B}-1\right)
 \label{eq:delta_l}
\end{equation}

% Role: explain LR selection at a chosen BS through the two loss increments.
This decomposition lets us obtain the optimal LR for any chosen BS. At that BS, the second term in Equation~\ref{eq:delta_l} gives the minimum loss increase achievable by optimizing LR, while the first term gives the additional increase from using a different LR. The optimal LR therefore makes the first term zero. This minimum occurs at $\eta=\eta^*(B/B^*)^{\gamma_{\eta,B}}$, where the curvature in log LR is $2h$. Thus, $\gamma_{\eta,B}$ determines how the optimal LR changes with BS, while $h$ controls how sharply loss rises around that LR.

% Role: explain the remaining BS dependence after optimizing LR.
With LR optimized at each BS, the second term describes how the minimum loss varies with BS. Its curvature in log BS at $B^*$ is $2h\rho$, so $\rho>0$ is the ratio of this curvature to the curvature $2h$ in log LR. We fit one $\rho$ per baseline ladder. Appendix~\ref{app:lrbs_term} derives the optimal LR at a fixed BS and the curvature expressions.

% Role: summarize the LR--BS parameters after defining all components.
Table~\ref{tab:response_parameters} summarizes the power laws and shape parameters of the LR--BS term for each baseline ladder.
\input{tables/response_parameters.tex}

\subsection{Loss prediction and hyperparameter selection}
\label{sec:fit_quality}
% Role: connect loss prediction on the ladders and at larger budgets to the use of predicted optima.
We evaluate loss predictions across the scaling ladder and at larger compute budgets, then examine how the predicted optima can guide LR--BS searches.

\paragraph{Fit quality across the scaling ladder.}
% Role: prediction comparisons and error definition.
We compare the base-surface predictions with each model--budget group's lowest measured loss (Figure~\ref{fig:isoflops}) and the full-law predictions with the losses of all swept runs. We report mean absolute percentage error (MAPE) at the group minima (Best-point) and over all swept runs (All-point), as defined in Appendix~\ref{app:error_metrics}.

% Role: the measured reason for the coupled base surface.
We first compare the coupled surface with the additive ablation, which fixes $\gamma_{MD}=1$ and is fitted to the same group minima. For the fitted range $0<\gamma_{MD}<1$, the coupled form makes the loss reductions from model scale and training data reinforce each other; the additive form keeps them independent (Appendix~\ref{app:additive}). Both forms have fitted IsoFLOP minima within the swept range of model sizes, but the additive form deviates most from the measured losses at the smallest and largest budgets (Figure~\ref{fig:isoflops}, bottom row). Its best-point MAPE is 0.48\% for Qwen3.5 and 0.62\% for OLMo3, compared with 0.13\% and 0.11\% for the coupled surface (Table~\ref{tab:additive_parameters}). The coupled surface therefore fits the measured IsoFLOP minima more closely across the scaling ladder.

\input{figures/isoflops_comparison_figure.tex}

% Role: full-law fit on all swept runs.
We then evaluate the full scaling laws, which fit the measured losses of all swept runs with a MAPE of 0.25\% for Qwen3.5 and 0.27\% for OLMo3 (Table~\ref{tab:baseline_summary}). Residuals show no trend across budgets, so the laws do not systematically over- or underpredict loss at any budget (Appendix~\ref{app:fit_quality}).

\input{tables/baseline_summary.tex}

% Role: compare the fitted LR--BS contours with measured loss differences.
We also compare the fitted LR--BS term with the measured loss differences in every model--budget group of the Qwen3.5 ladder (Figure~\ref{fig:response_maps}); Figure~\ref{fig:response_maps_olmo} in Appendix~\ref{app:lrbs_maps} shows the corresponding OLMo3 results. The measured loss differences follow the fitted contours, including the tilt that couples the optimal LR to the BS.

\input{figures/response_maps_figure.tex}

\paragraph{Loss prediction at larger compute budgets.}
% Role: extrapolation runs.
We next test loss predictions at larger budgets by freezing the fitted coefficients and training models using the ladder training setup and validation data (Figure~\ref{fig:high_compute} and Table~\ref{tab:high_compute}). These runs use the largest swept models and new 32-layer models, trained near the compute-optimal combination of model size and training token count at up to about ten times the largest swept budget.

\input{figures/high_compute_figure.tex}
\input{tables/high_compute_validation.tex}

% Role: extrapolation results.
The best measured loss at each swept budget lies within 0.5\% of the compute-optimal curve of $L^*$, and the extrapolation losses follow this curve beyond the swept range (Figure~\ref{fig:high_compute}). At the LR and BS used in training, the frozen scaling laws also predict the extrapolation losses within 0.5\%, with a MAPE of 0.32\% (Table~\ref{tab:high_compute}), comparable to the MAPE on the swept runs. Appendix~\ref{app:extrapolation_recipes} gives the training recipes and compares predictions from $L^*$ alone with those from the full scaling law.

\paragraph{Using predicted optima to guide LR--BS searches.}
% Role: compare the predicted optima with measured recipes and explain their use in Setting 3.
Finally, we compare the predicted optimal LR and BS with the best measured configuration in each model--budget group (Table~\ref{tab:lrbs_forms}). The predicted optimal LR, however, differs from the best measured LR with a MAPE of 35.1\% for Qwen3.5 and 23.1\% for OLMo3. The Setting~3 search should therefore be initialized with these predicted optima, and the final recipe should be selected using the measured losses. Appendices~\ref{app:lrbs_form} and~\ref{app:per_group} examine how alternative LR--BS forms and methods for estimating the optima affect prediction accuracy.

% Role: extend the use of predicted optima to Phase II and explain why longer training needs a direct search.
We also use the predicted optima to initialize the LR--BS search for the longer pretraining in Phase~II (Figure~\ref{fig:coverage}). To compare training length across model sizes, we use tokens per parameter (TPP), $D/N$, where $N$ is the parameter count excluding the input embedding. The extrapolation runs follow the fitted compute-optimal path to larger models and budgets while remaining in the TPP range of the ladders. In contrast, Phase~II pretraining uses about 130--160 TPP, compared with 10--14 TPP in the longest ladder runs at the Phase~II model scales. This combination of model scale and token exposure lies outside the swept range, so we search over LR and BS directly in Phase~II (Section~\ref{sec:phase2_baselines}).
\input{figures/coverage_figure.tex}

\subsection{Comparison with scaling laws from prior work}
\label{sec:literature}
% Role: purpose and base-surface comparison.
We next compare the coupled and additive base surfaces within each study that reports both (Table~\ref{tab:literature_loss}). In our baseline ladders and in prior work, the coupled form matches the measured losses more closely than the additive form. The fitted coupling exponents are $\gamma_{MD}=0.51$ for Qwen3.5 and 0.40 for OLMo3, compared with 0.31--0.41 for Skaling and 0.41 for OpenEuroLLM.

\input{tables/literature_loss.tex}

% Role: optimal LR and BS against laws from prior work.
We also compare laws for optimal LR and BS, evaluating each law from prior work on the path for which it is defined (Table~\ref{tab:literature_hp} and Figure~\ref{fig:literature_hp}). At a fixed model scale, optimal BS follows a power law in data size with fitted exponents from 0.46 to 0.57 (Table~\ref{tab:literature_hp}), close to the $D^{1/2}$ prediction derived from the Muon-type proxy~\citep{shulgin2026} in Appendix~\ref{app:lrbs_term}. The fitted power laws for $B^*$ also have a negative exponent in $M$ for each baseline ladder (Table~\ref{tab:response_parameters}); the optimal BS therefore depends on both $M$ and $D$ (Table~\ref{tab:design_choices}). By contrast, the optimal-LR exponents vary more across studies, whose optimizers, schedules, and parameterizations differ. We therefore fit a separate LR law to each baseline ladder.

\input{figures/literature_hp_figure.tex}
\input{tables/literature_hp.tex}

\begin{samepage}
\subsection{Implications for the validation framework}
\label{sec:implications}
% Role: how each setting reuses the ladders.
The settings of Section~\ref{sec:phase1} reuse the baseline ladders in two ways. Settings~1 and~2 use the best measured recipes of the nearest model--budget groups (Appendix~\ref{app:recipes}), and Setting~3 starts its LR--BS search from the predicted optima (Appendix~\ref{app:predicted_optima}). At a BS imposed by hardware, the search starts from the predicted optimal LR at that BS, given by Equation~\ref{eq:conditional_lr} in Appendix~\ref{app:lrbs_term}.

% Role: predicted optima initialize direct tuning in Phase II.
Phase~II pretraining uses much higher token counts at the 1B scale (Section~\ref{sec:fit_quality}). Its direct LR--BS search also starts from the predicted optima (Section~\ref{sec:phase2_baselines}).
\end{samepage}

%% file: tables/coupled_parameters.tex
\begin{table}[tp]
\caption{Coefficients of the base surface $L^*$ (Equation~\ref{eq:coupled}), fitted to the lowest measured loss of each model--budget group. $M$ is in FLOPs per token and $D$ in tokens.}
\label{tab:coupled_parameters}
\centering\footnotesize
\input{tables/generated/coupled_parameters.tex}
\end{table}

%% file: tables/generated/coupled_parameters.tex
% Generated by scripts/make_baseline_figures.py from frozen evidence; do not edit by hand.
\begin{tabular}{lrrrrrr}
\toprule
Baseline & $L_\infty$ & $\alpha_M$ & $\gamma_M$ & $\alpha_D$ & $\gamma_D$ & $\gamma_{MD}$\\
\midrule
Qwen3.5 & 1.722 & $6.77\times10^{4}$ & 0.520 & $1.90\times10^{4}$ & 0.453 & 0.505\\
OLMo3 & 1.488 & $3.96\times10^{5}$ & 0.576 & $5.99\times10^{4}$ & 0.475 & 0.402\\
\bottomrule
\end{tabular}

%% file: tables/response_parameters.tex
\begin{table}[tp]
\caption{Power laws and shape parameters of the LR--BS term (Equations~\ref{eq:power_laws} and~\ref{eq:delta_l}). The exponent $\gamma_{\eta,B}$ is fixed at $1/2$. $M$ is in FLOPs per token; $D$ and $B^*$ are in tokens.}
\label{tab:response_parameters}
\centering\small
\input{tables/generated/response_parameters.tex}
\end{table}

%% file: tables/generated/response_parameters.tex
% Generated by scripts/make_baseline_figures.py from frozen evidence; do not edit by hand.
\begin{tabular}{lcc}
\toprule
Parameter & Qwen3.5 & OLMo3\\
\midrule
$\eta^*$ & $421M^{-0.408}D^{-0.152}$ & $269M^{-0.389}D^{-0.154}$\\
$B^*$ & $36.8M^{-0.047}D^{+0.485}$ & $47.3M^{-0.036}D^{+0.457}$\\
$h$ & $17.9M^{-0.075}D^{-0.242}$ & $14.0M^{-0.097}D^{-0.203}$\\
$(\gamma_{\eta,B},\rho)$ & $(0.5,1.16)$ & $(0.5,1.15)$\\
\bottomrule
\end{tabular}

%% file: figures/isoflops_comparison_figure.tex
\begin{figure}[tp]
\centering
\includegraphics[width=\linewidth]{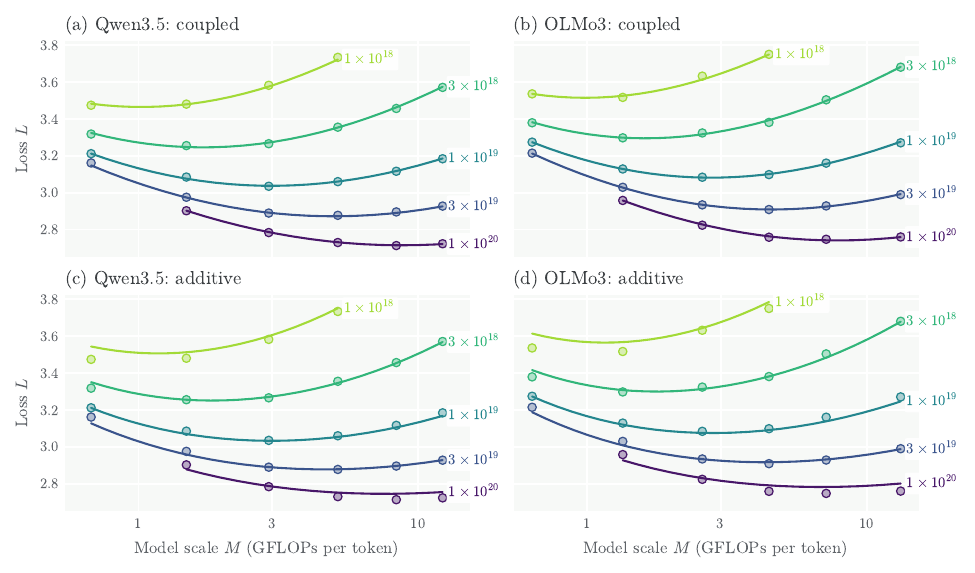}
\caption{Coupled and additive base surfaces for each baseline ladder. Circles mark the lowest measured loss per model--budget group; curves show the surfaces $L^*(M,D)$ fitted to these losses. Table~\ref{tab:coupled_parameters} gives the coupled-surface coefficients, Table~\ref{tab:additive_parameters} compares fit errors, and Table~\ref{tab:ladder_recipes} lists the measured minima.}
\label{fig:isoflops}
\label{fig:additive_curves}
\end{figure}

%% file: tables/baseline_summary.tex
\begin{table}[tp]
\caption{Loss-prediction errors of the full fixed-slope scaling laws on the swept runs. Mean absolute percentage error (MAPE) is reported at group minima (Best-point) and over all runs (All-point); P95 is the 95th percentile of absolute percentage error over all runs.}
\label{tab:baseline_summary}
\centering\small
\input{tables/generated/baseline_summary.tex}
\end{table}

%% file: tables/generated/baseline_summary.tex
% Generated by scripts/make_baseline_figures.py from frozen evidence; do not edit by hand.
\begin{tabular}{lrrr}
\toprule
Baseline & Best-point MAPE & All-point MAPE & P95\\
\midrule
Qwen3.5 & 0.22\% & 0.25\% & 0.80\%\\
OLMo3 & 0.17\% & 0.27\% & 0.69\%\\
\bottomrule
\end{tabular}

%% file: figures/response_maps_figure.tex
\begin{figure}[tp]
\centering
\includegraphics[width=\linewidth]{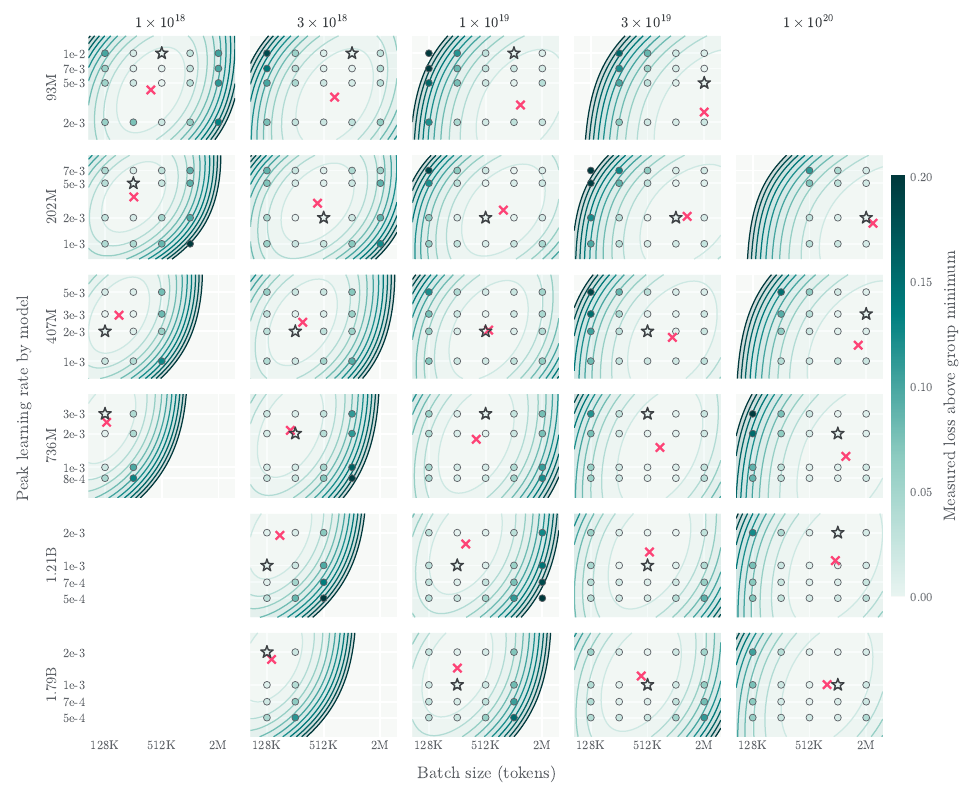}
\caption{LR--BS dependence of loss in the Qwen3.5 baseline ladder. Colors show the measured loss above each group minimum, and contours show the fitted LR--BS term $\Delta L$ at intervals of 0.02. Crosses mark predicted optima and stars the best measured configurations; empty panels are outside the sweep. Table~\ref{tab:response_parameters} gives the fitted laws.}
\label{fig:response_maps}
\end{figure}

%% file: figures/high_compute_figure.tex
\begin{figure}[tp]
\centering
\includegraphics[width=\linewidth]{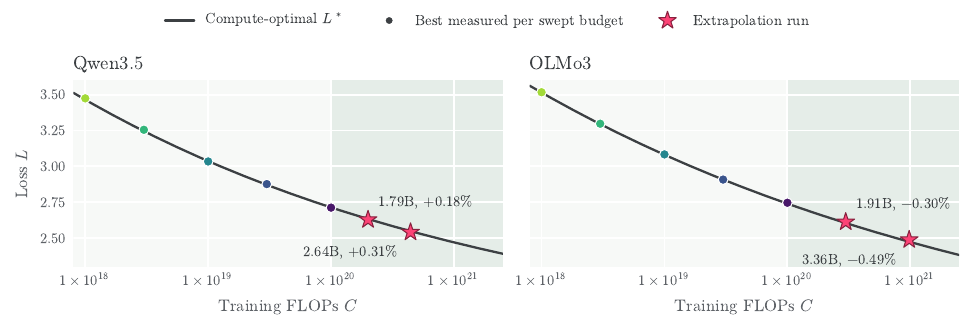}
\caption{Measured extrapolation losses compared with the compute-optimal loss $\min_{MD=C} L^*(M,D)$. Labels give the relative prediction errors $\widehat L/L-1$ of the full scaling law at each run's LR and BS, with the coefficients held fixed (Table~\ref{tab:high_compute}).}
\label{fig:high_compute}
\end{figure}

%% file: tables/high_compute_validation.tex
\begin{table}[tp]
\caption{Extrapolation runs beyond the largest swept budget, $1\times10^{20}$ FLOPs. Predictions use the frozen coefficients and each run's LR and BS. Relative error is $\widehat L/L-1$. Daggers mark model sizes beyond the swept range. Table~\ref{tab:extrapolation_recipes} gives the recipes, and Figure~\ref{fig:high_compute} compares the measured losses with the compute-optimal curve.}
\label{tab:high_compute}
\centering\footnotesize
\input{tables/generated/high_compute_validation.tex}
\end{table}

%% file: tables/generated/high_compute_validation.tex
% Generated by scripts/make_baseline_figures.py from frozen evidence; do not edit by hand.
\begin{tabular}{llrrrrr}
\toprule
Baseline & Model & FLOPs & TPP & $L$ & $\widehat L$ & Error\\
\midrule
Qwen3.5 & 1.79B & $2.00\times10^{20}$ & 9.1 & 2.6290 & 2.6337 & $+0.18\%$\\
Qwen3.5 & 2.64B$^\dagger$ & $4.43\times10^{20}$ & 9.4 & 2.5418 & 2.5498 & $+0.31\%$\\
OLMo3 & 1.91B & $2.98\times10^{20}$ & 11.8 & 2.6131 & 2.6053 & $-0.30\%$\\
OLMo3 & 3.36B$^\dagger$ & $9.80\times10^{20}$ & 12.9 & 2.4888 & 2.4766 & $-0.49\%$\\
\midrule
\multicolumn{6}{l}{Mean absolute percentage error} & 0.32\%\\
\bottomrule
\end{tabular}

%% file: figures/coverage_figure.tex
\begin{figure}[tp]
\centering
\includegraphics[width=\linewidth]{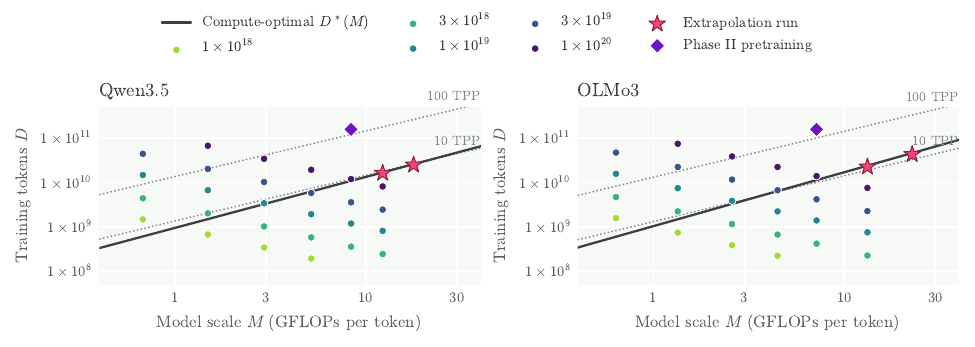}
\caption{Model scale and token counts in the baseline ladders, extrapolation runs (Section~\ref{sec:fit_quality}), and Phase~II pretraining (Section~\ref{sec:phase2_baselines}). Solid curves show the fitted compute-optimal paths $D^*(M)$; dotted lines indicate approximately 10 and 100 tokens per parameter (TPP).}
\label{fig:coverage}
\end{figure}

%% file: tables/literature_loss.tex
\begin{table}[tp]
\caption{Within-study comparisons of the coupled form (Equation~\ref{eq:coupled}) and the additive Chinchilla form~\citep{hoffmann2022}. The table denotes each study's outer exponent by $\gamma_{ND}$, corresponding to $\gamma_{MD}$ in Equation~\ref{eq:coupled}. SK is the Skaling study's own grid of training runs; RMSE denotes root mean square error. Prediction errors are evaluated on runs excluded from fitting in prior studies and at the swept group minima in this report.}
\label{tab:literature_loss}
\centering\footnotesize
\input{tables/generated/literature_loss.tex}
\end{table}

%% file: tables/generated/literature_loss.tex
% Generated by scripts/make_baseline_figures.py from frozen evidence; do not edit by hand.
\begin{tabular}{llrlrr}
\toprule
Study & Data & $\gamma_{ND}$ & Check & Coupled & Additive\\
\midrule
Skaling~\citep{skaling2026} & Farseer grid & 0.41 & Larger $N$, $D$; MAPE & 2.31\% & 2.46\%\\
Skaling~\citep{skaling2026} & SK grid & 0.31 & Larger $N$, $D$; MAPE & 0.70\% & 5.17\%\\
OpenEuroLLM~\citep{openeuro2026} & OpenEuroLLM ladder & 0.41 & Largest model; RMSE & 0.0056 & 0.0174\\
This report & Qwen3.5 & 0.51 & Group minima; MAPE & 0.13\% & 0.48\%\\
This report & OLMo3 & 0.40 & Group minima; MAPE & 0.11\% & 0.62\%\\
\bottomrule
\end{tabular}

%% file: figures/literature_hp_figure.tex
\begin{figure}[tp]
\centering
\includegraphics[width=\linewidth]{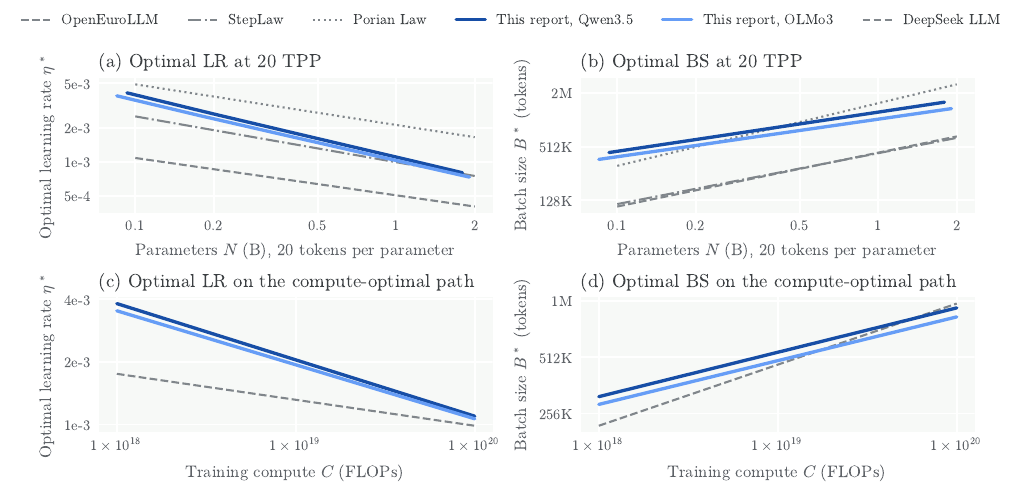}
\caption{Baseline scaling laws for optimal LR and BS compared with laws from prior work. Panels (a, b) compare the laws at 20 TPP with OpenEuroLLM~\citep{openeuro2026}, StepLaw~\citep{steplaw2025}, and Porian Law~\citep{porian2024}; panels (c, d) compare the laws with those of DeepSeek LLM~\citep{deepseekllm2024} at compute-optimal model sizes and training token counts. Baseline curves span the swept sizes and budgets; curves for prior work use the reported coefficients under each study's parameter-count convention. Table~\ref{tab:literature_hp} gives the exponents.}
\label{fig:literature_hp}
\end{figure}

%% file: tables/literature_hp.tex
\begin{table}[tp]
\caption{Scaling exponents of optimal LR and BS, measured as log--log slopes. The row blocks show exponents for a fixed model with varying data, for models of different sizes at 20 TPP, and for compute-optimal combinations of model size and training token count, respectively. When $N$ varies, we evaluate the baseline laws at the swept model sizes and the laws from prior work at 0.1B--2B parameters under each study's parameter-count convention; when $C$ varies, slopes are measured over $1\times10^{18}$--$1\times10^{20}$ FLOPs. The proxy row is theoretical; the other rows describe fitted laws. Figure~\ref{fig:literature_hp} plots the laws as $N$ or $C$ varies.}
\label{tab:literature_hp}
\centering\footnotesize
\input{tables/generated/literature_hp.tex}
\end{table}

%% file: tables/generated/literature_hp.tex
% Generated by scripts/make_baseline_figures.py from frozen evidence; do not edit by hand.
\begin{tabular}{llrr}
\toprule
Path & Law & $\eta^*$ exponent & $B^*$ exponent\\
\midrule
Fixed model, varying $D$ & \citet{shulgin2026} (proxy) & $-0.250$ & $+0.500$\\
 & OpenEuroLLM~\citep{openeuro2026} & $+0.086$ & $+0.460$\\
 & StepLaw~\citep{steplaw2025} & $+0.307$ & $+0.571$\\
 & This report, Qwen3.5 & $-0.152$ & $+0.485$\\
 & This report, OLMo3 & $-0.154$ & $+0.457$\\
\addlinespace
20 TPP, varying $N$ & OpenEuroLLM~\citep{openeuro2026} & $-0.331$ & $+0.605$\\
 & StepLaw~\citep{steplaw2025} & $-0.406$ & $+0.571$\\
 & Porian Law~\citep{porian2024} & $-0.360$ & $+0.703$\\
 & This report, Qwen3.5 & $-0.549$ & $+0.440$\\
 & This report, OLMo3 & $-0.531$ & $+0.422$\\
\addlinespace
Compute-optimal, varying $C$ & DeepSeek LLM~\citep{deepseekllm2024} & $-0.125$ & $+0.327$\\
 & This report, Qwen3.5 & $-0.271$ & $+0.238$\\
 & This report, OLMo3 & $-0.260$ & $+0.234$\\
\bottomrule
\end{tabular}

%% file: sections/05_long_training.tex
\section{Phase II: 1B Baselines and Training Recipe}
\label{sec:phase2_baselines}
% Role: why longer training is the second validation step and what the section provides.
To provide reusable baselines for the longer training and downstream evaluation in Phase~II (Section~\ref{sec:phase2}), we train 1B Qwen3.5 and OLMo3 models on 150B tokens at 8K context, followed by 10B tokens of long-context mid-training at 64K. Both baselines use the same data, token counts, and schedules within each training stage. We tune LR and BS for Qwen3.5, then reuse the selected baseline recipe for OLMo3 without tuning.

% Role: selection objectives and access to the supporting evidence.
Validation loss and downstream task scores favor different recipes. We therefore select recipes from the Pareto frontier using validation loss and the Base Main score for pretraining, and long-context scores and the Base Main score for mid-training. Section~\ref{sec:baseline_capabilities} reports the baselines' downstream evaluation results; Table~\ref{tab:stage1_sweep} reports the complete pretraining sweep; Appendix~\ref{app:phase2} gives the configurations and detailed analyses, including the complete mid-training sweep.

\subsection{Pretraining beyond the compute-optimal ladder}
\label{sec:pretraining}
% Role: regime change and the predicted starting point.
The Qwen3.5 pretraining run reaches about 130 TPP, compared with about 10 TPP in the longest ladder run at this model scale. We tune LR and BS for this longer training, starting from the predicted optima of the frozen scaling law (Table~\ref{tab:phase2_predictions}). At BS 4M, the predicted optimal LR is about 8e-4. Table~\ref{tab:stage1_sweep} gives the complete sweep, with loss predictions and Base Main scores.

\input{tables/phase2_predictions.tex}

% Role: measured recipe, its selection, and why completed schedules decide.
The final loss of every pretraining run is within 1\% of the lowest measured loss, yet the lowest loss and the highest Base Main score occur at different recipes. The selected recipe uses LR 3e-3 and BS 4M (Table~\ref{tab:stage1_sweep}); its LR is about 3.6 times the predicted optimum at the same BS. This choice relies on results from the final checkpoints of completed schedules: lower-LR runs have lower loss during the stable phase, but the run with the selected higher-LR recipe has the lowest loss after the final decay (Appendix~\ref{app:rank_reversal}).

\input{tables/stage1_sweep.tex}

% Role: loss predictions for both baselines, tested beyond the ladders.
The pretraining runs also test predictions outside the fitting regime: they extend beyond their baseline ladders in TPP and use a different warmup schedule (Appendix~\ref{app:pretraining_sweep}). The frozen Qwen3.5 scaling law overpredicts every pretraining loss, with a MAPE of about 1\% (Appendix~\ref{app:pretraining_sweep}). For OLMo3, the scaling law predicts the baseline loss within 0.6\% (Table~\ref{tab:phase2_predictions}).

\subsection{Long-context mid-training}
\label{sec:midtraining}
% Role: controlled mid-training design.
We then extend the pretrained model's context window to 64K through long-context mid-training. The Qwen3.5 sweep tests peak LR values from the final pretraining LR to the pretraining peak, using the pretraining BS of 4M and a doubled BS of 8M. Each run restores the model and optimizer state of the selected pretrained checkpoint and trains on the same 10B tokens.

% Role: mid-training selection and the resulting baseline recipe.
Using RULER as the primary development signal (Section~\ref{sec:phase2}), we select LR 1e-3 and BS 4M for mid-training (Figure~\ref{fig:midtraining_sweep} and Table~\ref{tab:midtraining_sweep}). At 64K, the length targeted by Phase~II, this recipe yields the highest RULER score: 57.0, compared with 51.9 when mid-training uses the final pretraining LR (Table~\ref{tab:long_training_summary}). The sweep also shows that the LR yielding the highest measured score is higher at BS 8M than at BS 4M for both RULER and HELMET at 64K (Appendix~\ref{app:midtraining}). The baseline recipe combines the selected settings for pretraining and mid-training: peak LR 3e-3 with BS 4M for pretraining, followed by 1e-3 with BS 4M for long-context mid-training (Table~\ref{tab:baseline_recipe}).

\input{figures/midtraining_sweep_figure.tex}
\input{tables/long_training_summary.tex}

% Role: cost of establishing the recipe and of each reuse.
The Qwen3.5 search across both training stages costs about $1.6\times10^{22}$ FLOPs. By comparison, reusing the selected recipe for a single run of both stages costs about $1.47\times10^{21}$ FLOPs for Qwen3.5 and $1.24\times10^{21}$ FLOPs for OLMo3, each less than a tenth of the search cost.

\subsection{Baseline capability results}
\label{sec:baseline_capabilities}
% Role: evaluation scope and the comparison conditions of the two baselines.
We next present downstream evaluation results for the pretrained and mid-trained checkpoints of both baselines on the Base Main and Base Long-Context suites of OLMo~3~\citep{olmo3}. Qwen3.5 uses the selected recipe, and OLMo3 reuses it without tuning. OLMo3 uses about 16\% less compute across the two training stages and a different inference implementation. The analysis therefore focuses on changes across training stages within each baseline.

% Role: score aggregation and evaluation context.
The Base Main score gives equal weight to Math, multiple-choice STEM (MC-STEM), multiple-choice non-STEM (MC-NonSTEM), and generative QA (GenQA); code tasks are excluded. Within each family, task and subtask scores are averaged with equal weights at each level according to the OLMES hierarchy. Base Main evaluation uses an input limit of 8K after pretraining and 64K after mid-training.

% Role: principal Base Main results and the task-level variation that conditions them.
After mid-training, the Base Main score rises from 27.9 to 46.3 for Qwen3.5 and from 26.2 to 40.5 for OLMo3 (Table~\ref{tab:base_main_tasks}). All four family averages increase. Within GenQA, however, HellaSwag, Jeopardy, and LAMBADA scores decrease for each baseline.

\input{tables/base_main_tasks.tex}

% Role: comparable long-context measurements at each evaluation length.
For the Base Long-Context suite, we also report RULER and HELMET scores at each evaluation length, using the same tasks, examples, and input limits across checkpoints (Table~\ref{tab:base_long_context}). The pretrained scores are reference measurements rather than selection criteria. RULER averages the scores from its 13 tasks with equal weights. Our HELMET~\citep{helmet} score gives equal weight to five categories: synthetic recall, retrieval-augmented generation, passage re-ranking, many-shot in-context learning, and long-document QA. Each category score averages its task scores equally; tasks scored by a model judge are excluded.

\input{tables/base_long_context.tex}

% Role: gains at the target evaluation length and what they establish for Phase~II.
At 64K, Qwen3.5's RULER score rises from 17.5 to 57.0 and its HELMET score from 23.3 to 39.1; OLMo3's scores rise from 2.7 to 36.1 and from 7.4 to 29.6. Both baselines achieve 64K long-context capability with the same two-stage recipe, tuned for Qwen3.5 and reused without tuning for OLMo3. Researchers can reuse the checkpoints, baseline recipe, and Base Main and Base Long-Context results in subsequent Phase~II comparisons.

%% file: tables/phase2_predictions.tex
\begin{table}[tp]
\caption{Predicted optima and losses for Phase~II pretraining from the frozen scaling laws. $\eta^*$ and $B^*$ are the predicted optimal LR and BS; the conditional LR is evaluated at BS 4M (Equation~\ref{eq:conditional_lr}). $L$ and $\widehat L$ are the measured and predicted losses for runs using the baseline recipe in Table~\ref{tab:baseline_recipe}; relative error is $\widehat L/L-1$.}
\label{tab:phase2_predictions}
\centering\small
\input{tables/generated/phase2_predictions.tex}
\end{table}

%% file: tables/generated/phase2_predictions.tex
% Generated by scripts/make_long_training_figures.py from frozen evidence; do not edit by hand.
\begin{tabular}{llrrrrrrr}
\toprule
Baseline & Model & TPP & $\eta^*$ & $B^*$ & $\eta^*(\cdot\mid B{=}\mathrm{4M})$ & $L$ & $\widehat L$ & Error\\
\midrule
Qwen3.5 & 1.21B & 129.8 & 7.49e-4 & 3.29M & 8.26e-4 & 2.4878 & 2.5256 & $+1.52\%$\\
OLMo3 & 1.00B & 156.9 & 7.5e-4 & 2.61M & 9.28e-4 & 2.5374 & 2.5516 & $+0.56\%$\\
\bottomrule
\end{tabular}

%% file: tables/stage1_sweep.tex
\begin{table}[tp]
\caption{Qwen3.5 1B pretraining sweep on about 150B tokens: measured loss and Base Main scores, with predictions $\widehat L$ from the frozen scaling law. Error is $\widehat L/L-1$. MC denotes multiple choice; GenQA denotes generative question answering.}
\label{tab:stage1_sweep}
\centering\footnotesize
\input{tables/generated/stage1_sweep.tex}
\end{table}

%% file: tables/generated/stage1_sweep.tex
% Generated by scripts/make_long_training_figures.py from frozen evidence; do not edit by hand.
\begin{tabular}{lrrrrrrrrr}
\toprule
& & & & & \multicolumn{5}{c}{Base Main}\\
\cmidrule(lr){6-10}$\eta$ & $B$ & $L$ & $\widehat L$ & Error & Score & Math & MC-STEM & MC-NonSTEM & GenQA\\
\midrule
8e-4 & 2M & 2.5004 & 2.5159 & $+0.62\%$ & 27.3 & 2.6 & 26.0 & 28.8 & 51.8\\
1e-3 & 2M & 2.5088 & 2.5171 & $+0.33\%$ & 26.9 & 2.6 & 24.2 & 29.0 & 52.0\\
2e-3 & 2M & 2.4958 & 2.5260 & $+1.21\%$ & 27.6 & 2.9 & 25.6 & \textbf{29.5} & 52.3\\
8e-4 & 4M & 2.5120 & 2.5138 & $+0.07\%$ & 27.1 & 2.5 & 26.7 & 28.6 & 50.5\\
1e-3 & 4M & 2.5055 & 2.5140 & $+0.34\%$ & 27.1 & 2.6 & 26.4 & 28.0 & 51.6\\
2e-3 & 4M & 2.4933 & 2.5190 & $+1.03\%$ & 27.4 & 2.7 & 25.7 & 28.8 & 52.3\\
3e-3 & 4M & \textbf{2.4878} & 2.5256 & $+1.52\%$ & 27.9 & \textbf{3.2} & 27.0 & 28.8 & \textbf{52.6}\\
4e-3 & 4M & 2.4941 & 2.5328 & $+1.55\%$ & \textbf{28.0} & 2.9 & \textbf{27.4} & 29.4 & 52.4\\
5e-3 & 4M & 2.4969 & 2.5400 & $+1.73\%$ & 27.3 & 2.9 & 25.4 & 28.3 & 52.5\\
3e-3 & 8M & 2.4985 & 2.5255 & $+1.08\%$ & 27.4 & 2.8 & 26.3 & 28.6 & 51.7\\
5e-3 & 8M & 2.4920 & 2.5352 & $+1.73\%$ & 27.5 & 2.8 & 26.7 & 28.0 & 52.4\\
\bottomrule
\end{tabular}

%% file: figures/midtraining_sweep_figure.tex
\begin{figure}[tp]
\centering
\includegraphics[width=\linewidth]{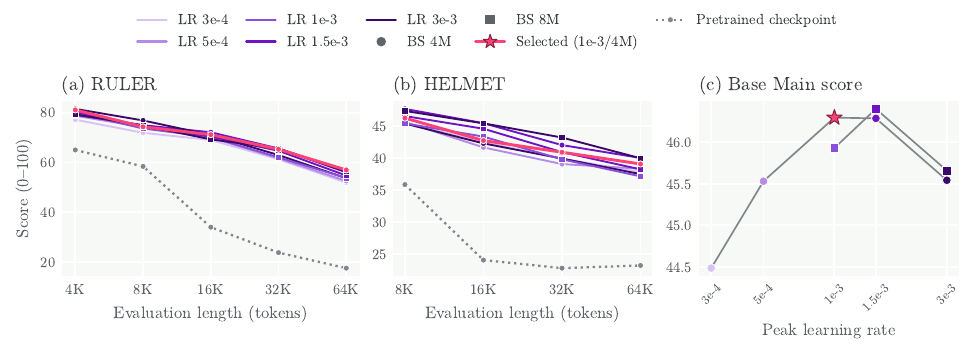}
\caption{Qwen3.5 long-context mid-training sweep: RULER (a) and HELMET (b) scores at each evaluation length, and the Base Main score against the peak LR (c). Base Main input limits are 8K for the pretrained checkpoint and 64K after mid-training. Tables~\ref{tab:midtraining_sweep} and~\ref{tab:context_lengths} give the full results.}
\label{fig:midtraining_sweep}
\end{figure}

%% file: tables/long_training_summary.tex
\begin{table}[tp]
\caption{64K RULER and HELMET scores before and after long-context mid-training at BS 4M. The Qwen3.5 rows compare the selected LR (1e-3), the final pretraining LR (3e-4), and the LR with the highest HELMET score (1.5e-3). OLMo3 reuses the baseline recipe without tuning. Tables~\ref{tab:midtraining_sweep} and~\ref{tab:olmo3_long} give the full sweep and the OLMo3 scores after each stage, respectively.}
\label{tab:long_training_summary}
\centering\small
\input{tables/generated/long_training_summary.tex}
\end{table}

%% file: tables/generated/long_training_summary.tex
% Generated by scripts/make_long_training_figures.py from frozen evidence; do not edit by hand.
\begin{tabular}{lrr}
\toprule
Checkpoint & RULER $\uparrow$ & HELMET $\uparrow$\\
\midrule
\multicolumn{3}{l}{\emph{Qwen3.5}}\\
Pretrained checkpoint & 17.5 & 23.3\\
Mid-trained at the final pretraining LR (3e-4) & 51.9 & 37.8\\
Mid-trained at the selected LR (1e-3) & 57.0 & 39.1\\
Highest HELMET (1.5e-3) & 56.0 & 40.0\\
\midrule
\multicolumn{3}{l}{\emph{OLMo3}}\\
Pretrained checkpoint & 2.7 & 7.4\\
Mid-trained with the baseline recipe (1e-3) & 36.1 & 29.6\\
\bottomrule
\end{tabular}

%% file: tables/base_main_tasks.tex
\begin{table}[tp]
\caption{Base Main task and family scores for both baselines before and after long-context mid-training. Evaluation uses input limits of 8K for the pretrained checkpoints and 64K after mid-training.}
\label{tab:base_main_tasks}
\centering\footnotesize
\input{tables/generated/base_main_tasks.tex}
\end{table}

%% file: tables/generated/base_main_tasks.tex
% Generated by scripts/make_long_training_figures.py from frozen evidence; do not edit by hand.
\begin{tabular}{lrrrr}
\toprule
& \multicolumn{2}{c}{Qwen3.5} & \multicolumn{2}{c}{OLMo3}\\
\cmidrule(lr){2-3}\cmidrule(lr){4-5}Task & Pretrained & Mid-trained & Pretrained & Mid-trained\\
\midrule
\textbf{Base Main score} & \textbf{27.9} & \textbf{46.3} & \textbf{26.2} & \textbf{40.5}\\
\addlinespace
\emph{Math} & \emph{3.2} & \emph{18.6} & \emph{2.1} & \emph{16.4}\\
\quad GSM8K & 5.3 & 32.0 & 3.1 & 29.1\\
\quad GSM Symbolic & 1.7 & 12.3 & 1.1 & 9.9\\
\quad MATH & 2.5 & 11.4 & 2.1 & 10.2\\
\addlinespace
\emph{MC-STEM} & \emph{27.0} & \emph{51.3} & \emph{25.2} & \emph{43.8}\\
\quad ARC & 26.5 & 67.2 & 24.6 & 54.7\\
\quad MedMCQA & 31.3 & 37.2 & 30.9 & 34.1\\
\quad MedQA & 23.3 & 29.9 & 21.4 & 25.2\\
\quad MMLU STEM & 25.7 & 42.3 & 22.8 & 36.2\\
\quad SciQ & 28.1 & 79.8 & 26.3 & 68.7\\
\addlinespace
\emph{MC-NonSTEM} & \emph{28.8} & \emph{59.3} & \emph{27.7} & \emph{49.2}\\
\quad CoQA (MC) & 26.6 & 77.5 & 27.2 & 61.2\\
\quad CSQA & 21.2 & 48.4 & 19.5 & 39.8\\
\quad DROP (MC) & 25.2 & 42.6 & 25.7 & 33.5\\
\quad Jeopardy (MC) & 24.6 & 60.9 & 26.2 & 47.3\\
\quad MMLU Humanities & 25.7 & 51.1 & 24.2 & 42.4\\
\quad MMLU Other & 28.3 & 52.1 & 25.1 & 46.3\\
\quad MMLU Social Sciences & 27.3 & 54.7 & 22.7 & 46.8\\
\quad Natural Questions (MC) & 25.7 & 54.6 & 24.4 & 44.1\\
\quad PIQA & 50.3 & 62.5 & 50.1 & 58.5\\
\quad Social IQa & 36.0 & 63.9 & 33.6 & 60.4\\
\quad SQuAD (MC) & 25.6 & 84.3 & 25.9 & 61.4\\
\addlinespace
\emph{GenQA} & \emph{52.6} & \emph{56.0} & \emph{49.8} & \emph{52.4}\\
\quad Basic Skills & 71.6 & 77.9 & 69.0 & 75.1\\
\quad CoQA & 55.9 & 59.3 & 54.3 & 55.2\\
\quad DROP & 25.9 & 41.0 & 23.6 & 35.6\\
\quad HellaSwag & 62.8 & 60.4 & 58.9 & 57.0\\
\quad Jeopardy & 32.7 & 30.1 & 27.6 & 24.4\\
\quad LAMBADA & 60.3 & 57.2 & 57.9 & 54.8\\
\quad Natural Questions & 12.9 & 15.1 & 13.0 & 13.7\\
\quad SQuAD & 77.6 & 88.2 & 72.1 & 83.9\\
\quad WinoGrande & 73.5 & 74.7 & 71.8 & 72.3\\
\bottomrule
\end{tabular}

%% file: tables/base_long_context.tex
\begin{table}[tp]
\caption{Base Long-Context suite results~\citep{olmo3}: RULER and HELMET before and after long-context mid-training, at each evaluation length. Tables~\ref{tab:context_lengths} and~\ref{tab:context_categories} give the full sweep and the 64K category scores, respectively.}
\label{tab:base_long_context}
\centering\footnotesize
\input{tables/generated/base_long_context.tex}
\end{table}

%% file: tables/generated/base_long_context.tex
% Generated by scripts/make_long_training_figures.py from frozen evidence; do not edit by hand.
\begin{tabular}{llrrrrrrrrr}
\toprule
& & \multicolumn{5}{c}{RULER} & \multicolumn{4}{c}{HELMET}\\
\cmidrule(lr){3-7}\cmidrule(lr){8-11}Baseline & Checkpoint & 4K & 8K & 16K & 32K & 64K & 8K & 16K & 32K & 64K\\
\midrule
Qwen3.5 & Pretrained & 65.1 & 58.4 & 34.0 & 23.8 & 17.5 & 35.9 & 24.1 & 22.9 & 23.3\\
Qwen3.5 & Mid-trained & 81.2 & 74.4 & 71.2 & 65.3 & 57.0 & 46.3 & 42.8 & 40.9 & 39.1\\
\addlinespace
OLMo3 & Pretrained & 69.4 & 51.9 & 7.9 & 4.1 & 2.7 & 34.6 & 8.1 & 9.1 & 7.4\\
OLMo3 & Mid-trained & 75.3 & 63.2 & 52.4 & 43.3 & 36.1 & 42.1 & 37.1 & 34.7 & 29.6\\
\bottomrule
\end{tabular}

%% file: sections/06_related_work.tex
\section{Related Work}
\label{sec:related}
% Role: position the public contribution process relative to reusable baselines and open training and evaluation resources.
\paragraph{Open research resources.}
Shared baselines and documented evaluations support reproducible model comparisons. OLMo~3~\citep{olmo3} releases training data, recipes, and intermediate checkpoints, while OLMES~\citep{olmes} standardizes evaluation choices that affect comparisons between models. Using OLMo~3 data and OLMES evaluation, \archspace{} connects reusable architecture baselines to a continuing public contribution process: proposals, implementations, experimental records, reasoning, and reviews remain linked for later citation and reuse (Section~\ref{sec:public_review}).

% Role: establish the sources of compute-optimal comparisons and consistent FLOPs accounting.
\paragraph{Scaling laws for architecture comparison.}
\citet{kaplan2020} established power-law relationships between language-model loss, model size, data, and compute. Chinchilla~\citep{hoffmann2022} used IsoFLOP profiles and an additive loss surface to determine compute-optimal model sizes and training token counts. DeepSeek LLM~\citep{deepseekllm2024} represented model scale by non-embedding FLOPs per token, while \citet{porian2024} showed how output-layer compute, warmup, and optimizer tuning explain discrepancies in compute-optimal scaling. These studies provide the basis for Phase~I's IsoFLOP comparisons with consistent accounting of training FLOPs (Section~\ref{sec:phase1}).

% Role: connect the closest loss-surface models to the base surface used in the manuscript.
The form of the loss surface determines how predictions vary across model sizes and training token counts. Skaling~\citep{skaling2026} couples the model and data terms through an outer exponent, reducing systematic residuals of the additive Chinchilla form. OpenEuroLLM~\citep{openeuro2026} fitted this coupled surface to the lowest measured loss of each model--budget group. We adopt the coupled form and this fitting target for the base surface $L^*(M,D)$, using training FLOPs per token $M$ as model scale, and add an LR--BS term to describe loss away from the predicted optima (Section~\ref{sec:analysis}).

% Role: distinguish the paths and variables of the compared hyperparameter laws and the way their optima are estimated.
\paragraph{Hyperparameter scaling and loss modeling.}
Hyperparameter scaling laws range from relationships along a prescribed training path to functions of model size and token count. DeepSeek LLM~\citep{deepseekllm2024} fits optimal LR and BS along the compute-optimal path, whereas the Porian Law~\citep{porian2024} describes a path at about 20 TPP. Microsoft Law~\citep{bjorck2025} models optimal LR as a product of powers of model size and token count at a fixed BS, estimating the optima from quadratic fits in log LR. StepLaw~\citep{steplaw2025} uses this product form for LR and a token-count-only power law for BS. OpenEuroLLM~\citep{openeuro2026} first estimates optima from a quadratic surface in the original LR and BS coordinates within each model--budget group, then fits product power laws for both LR and BS. We fit power laws for the optima through the losses of all swept runs and compare prior laws along the paths on which they are defined (Sections~\ref{sec:analysis} and~\ref{sec:literature}).

% Role: separate prior optimization proxies from our derivation and fitted LR--BS model.
Optimization theory provides a complementary basis for modeling the effect of LR and BS on loss. \citet{shulgin2026} derive a proxy for Muon-type optimization, and \citet{su2026scaling} analyzes generalized power-law proxies by minimizing first over LR and then over BS. Starting from the fixed-momentum Muon-type proxy, we derive its cosh representation in log space. This representation and the proxy's square-root scaling of optimal LR with BS motivate our fixed-slope LR--BS term, which we combine with fitted power laws for the optima and curvature (Section~\ref{sec:analysis}; Appendix~\ref{app:lrbs_term}).

% Role: position Phase II against direct architecture ablations and explain the complementary long-context evaluations.
\paragraph{Architecture evaluation after context extension.}
OlmPool~\citep{bertsch2026cracks} directly studies how architectural choices affect long-context extension, finding that short-context metrics poorly predict the resulting long-context capabilities. It trains 7B models on about 140B pretraining tokens followed by 10B long-context tokens. Phase~II adopts the same sequence of pretraining and long-context mid-training, using 1B models trained on 150B+10B tokens to evaluate architectural gains beyond the compute--loss comparisons of Phase~I (Section~\ref{sec:phase2}). RULER~\citep{ruler} tests retrieval, multi-hop tracing, aggregation, and question answering with synthetic tasks; HELMET~\citep{helmet} covers a broader range of downstream applications. Phase~II uses these benchmarks in the Base Long-Context suite and evaluates general capabilities with the Base Main suite (Section~\ref{sec:baseline_capabilities}).

%% file: sections/08_conclusion.tex
\section{Conclusion}
\label{sec:conclusion}
% Role: what the report provides: platform, framework, and initial references.
\archspace{} records, reviews, and references architecture experiments under a standardized validation framework with shared data and evaluation. In Phase~I, loss is compared across the compute budgets of the scaling ladder. Phase~II then evaluates downstream capabilities in 1B models after pretraining on 150B tokens and after long-context mid-training on 10B tokens. We provide Qwen3.5 and OLMo3 baseline ladders, fitted scaling laws, measured recipes, and Phase~II baselines as initial references for this workflow.

% Role: scaling-law accuracy and its use in choosing model size, training token count, and recipes.
The scaling laws combine a coupled base surface with a fixed-slope cosh LR--BS term and fit the measured losses from both baseline ladders with a MAPE of 0.25\%--0.27\%. The coupled base surface fits the group minima more accurately than its additive ablation, and the fitted coupling exponents and data exponents of optimal BS are close to those in prior work. The frozen laws predict extrapolation losses at up to ten times the largest swept budget with a MAPE of 0.32\%. These laws also give compute-optimal model sizes and training token counts. Locating optimal LR, however, is harder than predicting loss, so we use the predicted optima to initialize direct recipe searches.

% Role: the Phase~II search, capability gains, and recipe reuse.
In Phase~II, Qwen3.5's selected pretraining LR is about 3.6 times the predicted optimum at the same BS. Loss predictions nevertheless remain close to the measurements for both baselines. Long-context mid-training then raises the Base Main scores and establishes 64K long-context capability for both baselines. The recipe tuned for Qwen3.5 also yields large gains for OLMo3 when reused without tuning.

% Role: vision closing: the proposal-to-baseline cycle as the platform's aim.
With these baselines, \archspace{} offers the community an ongoing cycle of architecture research. Contributors introduce an idea through an Architecture Proposal, evaluate it under the validation framework, and submit the resulting record for public review. A Verified contribution can then become a new baseline for the next proposal. The reviewed records let later studies reproduce comparisons and reuse implementations. This cycle addresses the gaps identified at the start of this report: incomplete public records of the experiments and reasoning behind successful and unsuccessful architectural decisions, and research attention concentrated around industrial releases. \archspace{} aims to broaden the range of ideas that receive public review, citation, and recognition, making today's baselines the starting points for tomorrow's architecture research.

%% file: sections/10_appendix_configurations.tex
\section{Architecture Configurations}
\label{app:architectures}
% Role: what the table covers and the properties shared by both baselines.
Table~\ref{tab:exact_shapes} lists the Qwen3.5 and OLMo3 model shapes, parameter counts, and per-token training FLOPs. The baseline models use dense SwiGLU feed-forward blocks and untied input embeddings and output heads; all baseline series use the initialization described in Appendix~\ref{app:initialization}.

% Role: Qwen3.5 layer structure, attention heads, position encoding, and normalization.
Qwen3.5 repeats three Gated DeltaNet layers followed by one gated full-attention layer. Each full-attention head has dimension 256. The query/key--value head counts are 4/1, 6/1, 8/1, 10/2, 12/2, and 14/2 for the six swept sizes in ascending order, and 16/2 for the 2.64B extrapolation model. In Gated DeltaNet layers, the key/value head counts are 4/8, 6/12, 8/16, 10/20, 12/24, 14/28, and 16/32 in the same model order. The key and value heads each have dimension 128, and the convolution kernel has size 4. Qwen3.5 uses rotary position embeddings (RoPE) with base $1\times10^7$ and rotary fraction 0.25, and RMS normalization (RMSNorm) with $\epsilon=1\times10^{-6}$.

% Role: OLMo3 layer structure, attention heads, position encoding, and normalization.
OLMo3 repeats three sliding-window attention layers with a 4,096-token window followed by one full-attention layer. The numbers of query, key, and value heads are equal. Across the six swept sizes in ascending order, the head counts are 8, 8, 16, 16, 24, and 32, with corresponding head dimensions of 64, 96, 64, 80, 64, and 64. The 3.36B extrapolation model has 32 heads of dimension 80. OLMo3 uses RoPE with base 500,000, query--key normalization, and reordered RMSNorm with $\epsilon=1\times10^{-6}$.

\input{tables/architecture_shapes.tex}

%% file: tables/architecture_shapes.tex
\begin{table}[tp]
\caption{Model shapes, parameter counts, and per-token training FLOPs for the Qwen3.5 and OLMo3 baselines. $d$ and $d_{\mathrm{ff}}$ are model and feed-forward widths; $N$ excludes the input embedding and $N_{\mathrm{total}}$ includes it. $M$ is training FLOPs per token and includes the cost of attention at an 8K-token context length. Daggers mark models used only in the extrapolation runs of Section~\ref{sec:fit_quality}.}
\label{tab:exact_shapes}
\centering\footnotesize
\begin{tabular}{llrrrrrr}
\toprule
Baseline & Model & Layers & $d$ & $d_{\mathrm{ff}}$ & $N$ & $N_{\mathrm{total}}$ & $M$\\
\midrule
Qwen3.5 & 93M & 8 & 512 & 2048 & 92,907,872 & 144,288,096 & 676,930,560\\
 & 202M & 12 & 768 & 2560 & 202,411,608 & 279,481,944 & 1,483,310,592\\
 & 407M & 16 & 1024 & 3584 & 407,475,072 & 510,235,520 & 2,922,792,960\\
 & 736M & 20 & 1280 & 4608 & 735,791,320 & 864,241,880 & 5,161,536,000\\
 & 1.21B & 24 & 1536 & 5632 & 1,212,956,256 & 1,367,096,928 & 8,353,115,136\\
 & 1.79B & 28 & 1792 & 6144 & 1,794,704,408 & 1,974,535,192 & 12,231,937,536\\
 & 2.64B$^\dagger$ & 32 & 2048 & 7168 & 2,642,291,200 & 2,847,812,096 & 17,765,535,744\\
\addlinespace
OLMo3 & 85M & 8 & 512 & 2048 & 84,951,552 & 136,331,776 & 635,489,280\\
 & 176M & 12 & 768 & 2560 & 176,198,400 & 253,268,736 & 1,340,195,328\\
 & 346M & 16 & 1024 & 3584 & 346,096,640 & 448,857,088 & 2,579,699,712\\
 & 614M & 20 & 1280 & 4608 & 613,520,640 & 741,971,200 & 4,467,248,640\\
 & 1.00B & 24 & 1536 & 5632 & 1,003,636,224 & 1,157,776,896 & 7,153,837,056\\
 & 1.91B & 28 & 2048 & 7168 & 1,908,639,744 & 2,114,160,640 & 13,212,758,016\\
 & 3.36B$^\dagger$ & 32 & 2560 & 9216 & 3,361,016,320 & 3,617,917,440 & 22,681,697,280\\
\bottomrule
\end{tabular}
\end{table}

%% file: sections/11_appendix_baseline_recipes.tex
\section{Training Recipe and Best Measured Recipes}
\label{app:recipes}

\subsection{Optimizer and parameter groups}
\label{app:optimizer}
% Role: which parameters receive each update and regularization.
AdamW updates the input embeddings, the output head, and the scalar, vector, and convolution parameters. Muon optimizes all other two-dimensional matrix parameters, including projection and feed-forward matrices (Table~\ref{tab:optimizer_groups}). AdamW uses betas $(0.9,0.95)$ and gradient-norm clipping at 1.0; clipping is disabled for the Muon parameters.
\input{tables/optimizer_groups.tex}

% Role: explain orthogonalization and the shared LR convention.
Muon approximately orthogonalizes gradient momentum, preserving its singular-vector directions while bringing the singular values toward one~\citep{jordan2024muon}. We use momentum 0.9 and approximate the orthogonalization with five Newton--Schulz iterations using a fifth-degree polynomial, avoiding an explicit singular-value decomposition. Following the Muon update rule~\citep{liu2025muon}, the optimization part of the matrix update is
\begin{equation}
 \Delta W_t=-\eta_t\,0.2\sqrt{\max(d_{\mathrm{out}},d_{\mathrm{in}})}\,O_t,
 \label{eq:muon_update}
\end{equation}
where $O_t$ is the approximately orthogonalized momentum, $d_{\mathrm{out}}$ and $d_{\mathrm{in}}$ are the dimensions of the matrix being orthogonalized, and $\eta_t$ is the scheduled LR. The root mean square of $O_t$ varies with the matrix dimensions. The square-root factor compensates for this variation, and the factor 0.2 scales the update to a magnitude near that of AdamW. These factors let the two optimizers share $\eta_t$; the reported LR specifies the peak of this shared schedule.

\subsection{Initialization}
\label{app:initialization}
% Role: define the common width and residual-depth scales with precise attribution.
All baseline series use a scaled truncated-normal initialization adapted from a Megatron-based training recipe. The base initialization scale depends on model width as in \citeauthor{nguyen2019transformers}~\citep{nguyen2019transformers}; residual output projections are additionally scaled with depth. For model width $d$ and block count $n_{\mathrm{layers}}$, the scales are
\begin{equation}
 \sigma_{\mathrm{base}}=\sqrt{\frac{2}{5d}},\qquad
 \sigma_{\mathrm{out}}=\frac{\sigma_{\mathrm{base}}}{\sqrt{n_{\mathrm{layers}}}}.
 \label{eq:initialization}
\end{equation}
Embeddings, input projections, convolution weights, and the output head use a zero-mean normal distribution with scale $\sigma_{\mathrm{base}}$; residual output projections in the Gated DeltaNet and attention layers and in the feed-forward networks use $\sigma_{\mathrm{out}}$. Each normal distribution is truncated at three standard deviations on either side of zero. Finally, normalization weights are initialized to one.

% Role: initialize GDN's positive decay and time-step parameters.
Gated DeltaNet represents its decay rate with a logarithmic parameter so that the rate remains positive~\citep{yang2025gdn}. We initialize this parameter as $\log A$, with $A$ sampled uniformly between $1\times10^{-2}$ and $16$. The time-step bias is initialized by applying the inverse softplus to a value sampled log-uniformly between $1\times10^{-3}$ and $1\times10^{-1}$.

\subsection{Schedule and search grids}
\label{app:schedule_grids}
% Role: schedule and reuse of the stable phase across budgets.
The shared WSD schedule~\citep{hu2024minicpm} warms up the LR over 33.6M ($2^{25}$) tokens, holds it at the peak, and decays it over the final 20\% of updates to 10\% of the peak with a square-root profile~\citep{hagele2024}. The stable phase is independent of the total training length. We therefore reuse this phase across runs that differ only in budget, then apply a separate LR decay for each budget, reducing the cost of the sweeps. The Phase~II schedules are specified in Appendix~\ref{app:phase2_config}.

% Role: LR grids and their BS range across model--budget groups.
We choose LR values around the expected optimum, with lower values for deeper models (Table~\ref{tab:lr_grids}). This choice follows the decrease in optimal LR with model size reported by the Porian Law~\citep{porian2024} and StepLaw~\citep{steplaw2025}. The tested BS values range from 128K to 2M tokens; for each model--budget pair, they give 500--200,000 updates and keep warmup below 20\% of the training tokens.
\input{tables/lr_grids.tex}

% Role: how the measured recipes are used.
The baseline's Setting~1 curve plots the best measured loss at each budget (Table~\ref{tab:best_recipes}). Setting~1 and Setting~2 reuse the best measured LR and BS from the model--budget groups listed in Table~\ref{tab:ladder_recipes}. The predicted optima in Table~\ref{tab:optima} initialize the Setting~3 search.

\input{tables/best_recipes.tex}
\input{tables/ladder_recipes.tex}

%% file: tables/optimizer_groups.tex
\begin{table}[tp]
\caption{Optimizer assignments and weight decay for Qwen3.5 and OLMo3. Gated DeltaNet (GDN) rows apply only to Qwen3.5.}
\label{tab:optimizer_groups}
\centering\small
\input{tables/generated/optimizer_groups.tex}
\end{table}

%% file: tables/generated/optimizer_groups.tex
% Generated by scripts/make_baseline_figures.py from the shared training recipe; do not edit by hand.
\begin{tabular}{llr}
\toprule
Parameters & Optimizer & Weight decay\\
\midrule
Input embedding matrix & AdamW & 0\\
Output head matrix & AdamW & 0.1\\
\addlinespace
Normalization weights & AdamW & 0\\
GDN decay and time-step parameters & AdamW & 0\\
GDN depthwise convolution weights & AdamW & 0.1\\
\midrule
All other 2D matrix parameters & Muon & 0.1\\
\bottomrule
\end{tabular}

%% file: tables/lr_grids.tex
\begin{table}[tp]
\caption{Peak LR values tested in the Qwen3.5 and OLMo3 baseline ladders, grouped by model depth. Each LR value is tested at BS 128K, 256K, 512K, 1M, and 2M tokens where feasible.}
\label{tab:lr_grids}
\centering\small
\begin{tabular}{ll}
\toprule
Layers & $\eta$ grid\\
\midrule
8 & 2e-3, 5e-3, 7e-3, 1e-2\\
12 & 1e-3, 2e-3, 5e-3, 7e-3\\
16 & 1e-3, 2e-3, 3e-3, 5e-3\\
20 & 8e-4, 1e-3, 2e-3, 3e-3\\
24 and 28 & 5e-4, 7e-4, 1e-3, 2e-3\\
\bottomrule
\end{tabular}
\end{table}

%% file: tables/best_recipes.tex
\begin{table}[tp]
\caption{Best measured recipe of each baseline at each compute budget.}
\label{tab:best_recipes}
\centering\footnotesize
\input{tables/generated/best_recipes.tex}
\end{table}

%% file: tables/generated/best_recipes.tex
% Generated by scripts/make_baseline_figures.py from frozen evidence; do not edit by hand.
\begin{tabular}{lllrrrr}
\toprule
Baseline & FLOPs & Model & $D$ & $\eta$ & $B$ & $L$\\
\midrule
Qwen3.5 & $1\times10^{18}$ & 93M & 1.48B & 1e-2 & 512K & 3.4743\\
Qwen3.5 & $3\times10^{18}$ & 202M & 2.02B & 2e-3 & 512K & 3.2553\\
Qwen3.5 & $1\times10^{19}$ & 407M & 3.42B & 2e-3 & 512K & 3.0346\\
Qwen3.5 & $3\times10^{19}$ & 736M & 5.81B & 3e-3 & 512K & 2.8768\\
Qwen3.5 & $1\times10^{20}$ & 1.21B & 12.0B & 2e-3 & 1M & 2.7127\\
\addlinespace
OLMo3 & $1\times10^{18}$ & 176M & 746M & 5e-3 & 256K & 3.5170\\
OLMo3 & $3\times10^{18}$ & 176M & 2.24B & 2e-3 & 512K & 3.2975\\
OLMo3 & $1\times10^{19}$ & 346M & 3.88B & 2e-3 & 512K & 3.0837\\
OLMo3 & $3\times10^{19}$ & 614M & 6.72B & 2e-3 & 512K & 2.9083\\
OLMo3 & $1\times10^{20}$ & 1.00B & 14.0B & 1e-3 & 512K & 2.7468\\
\bottomrule
\end{tabular}

%% file: tables/ladder_recipes.tex
\begin{table}[tp]
\caption{Best measured recipe in each model--budget group.}
\label{tab:ladder_recipes}
\centering\footnotesize
\input{tables/generated/ladder_recipes.tex}
\end{table}

%% file: tables/generated/ladder_recipes.tex
% Generated by scripts/make_baseline_figures.py from frozen evidence; do not edit by hand.
\begin{tabular}{llrrrrlrrrr}
\toprule
& \multicolumn{5}{c}{Qwen3.5} & \multicolumn{5}{c}{OLMo3}\\
\cmidrule(lr){2-6}\cmidrule(lr){7-11}FLOPs & Model & $D$ & $\eta$ & $B$ & $L$ & Model & $D$ & $\eta$ & $B$ & $L$\\
\midrule
$1\times10^{18}$ & 93M & 1.48B & 1e-2 & 512K & 3.4743 & 85M & 1.57B & 5e-3 & 512K & 3.5361\\
 & 202M & 674M & 5e-3 & 256K & 3.4804 & 176M & 746M & 5e-3 & 256K & 3.5170\\
 & 407M & 342M & 2e-3 & 128K & 3.5826 & 346M & 388M & 5e-3 & 128K & 3.6323\\
 & 736M & 194M & 3e-3 & 128K & 3.7343 & 614M & 224M & 3e-3 & 128K & 3.7505\\
\midrule
$3\times10^{18}$ & 93M & 4.43B & 1e-2 & 1M & 3.3180 & 85M & 4.72B & 5e-3 & 512K & 3.3793\\
 & 202M & 2.02B & 2e-3 & 512K & 3.2553 & 176M & 2.24B & 2e-3 & 512K & 3.2975\\
 & 407M & 1.03B & 2e-3 & 256K & 3.2656 & 346M & 1.16B & 2e-3 & 256K & 3.3238\\
 & 736M & 581M & 2e-3 & 256K & 3.3553 & 614M & 672M & 2e-3 & 256K & 3.3811\\
 & 1.21B & 359M & 1e-3 & 128K & 3.4568 & 1.00B & 419M & 2e-3 & 128K & 3.5039\\
 & 1.79B & 245M & 2e-3 & 128K & 3.5713 & 1.91B & 227M & 1e-3 & 128K & 3.6810\\
\midrule
$1\times10^{19}$ & 93M & 14.8B & 1e-2 & 1M & 3.2115 & 85M & 15.7B & 5e-3 & 1M & 3.2741\\
 & 202M & 6.74B & 2e-3 & 512K & 3.0846 & 176M & 7.46B & 2e-3 & 512K & 3.1284\\
 & 407M & 3.42B & 2e-3 & 512K & 3.0346 & 346M & 3.88B & 2e-3 & 512K & 3.0837\\
 & 736M & 1.94B & 3e-3 & 512K & 3.0597 & 614M & 2.24B & 2e-3 & 512K & 3.0979\\
 & 1.21B & 1.20B & 1e-3 & 256K & 3.1161 & 1.00B & 1.40B & 2e-3 & 256K & 3.1603\\
 & 1.79B & 818M & 1e-3 & 256K & 3.1839 & 1.91B & 757M & 1e-3 & 256K & 3.2705\\
\midrule
$3\times10^{19}$ & 93M & 44.3B & 5e-3 & 2M & 3.1612 & 85M & 47.2B & 2e-3 & 1M & 3.2144\\
 & 202M & 20.2B & 2e-3 & 1M & 2.9750 & 176M & 22.4B & 2e-3 & 1M & 3.0290\\
 & 407M & 10.3B & 2e-3 & 512K & 2.8892 & 346M & 11.6B & 2e-3 & 512K & 2.9336\\
 & 736M & 5.81B & 3e-3 & 512K & 2.8768 & 614M & 6.72B & 2e-3 & 512K & 2.9083\\
 & 1.21B & 3.59B & 1e-3 & 512K & 2.8953 & 1.00B & 4.19B & 1e-3 & 512K & 2.9279\\
 & 1.79B & 2.45B & 1e-3 & 512K & 2.9271 & 1.91B & 2.27B & 1e-3 & 512K & 2.9893\\
\midrule
$1\times10^{20}$ & 202M & 67.4B & 2e-3 & 2M & 2.9017 & 176M & 74.6B & 2e-3 & 1M & 2.9575\\
 & 407M & 34.2B & 3e-3 & 2M & 2.7834 & 346M & 38.8B & 2e-3 & 1M & 2.8232\\
 & 736M & 19.4B & 2e-3 & 1M & 2.7292 & 614M & 22.4B & 2e-3 & 1M & 2.7579\\
 & 1.21B & 12.0B & 2e-3 & 1M & 2.7127 & 1.00B & 14.0B & 1e-3 & 512K & 2.7468\\
 & 1.79B & 8.18B & 1e-3 & 1M & 2.7226 & 1.91B & 7.57B & 1e-3 & 512K & 2.7593\\
\bottomrule
\end{tabular}

%% file: sections/12_appendix_scaling_laws.tex
\section{Scaling-Law Fitting and Validation}
\label{app:fitting}
% Role: what is fitted and on which measurements.
The scaling laws are fitted to validation losses measured at the final checkpoints after WSD decay. Validation loss is averaged with equal weights over eleven domains, as specified in Section~\ref{sec:phase1}: C4 English, Dolma books, Dolma common crawl, Dolma pes2o, Dolma Reddit, Dolma Stack, Dolma Wiki, ICE, M2D2 S2ORC, Pile, and WikiText-103~\citep{olmo3}.

\subsection{LR--BS term}
\label{app:lrbs_term}
% Role: motivate the optimization proxy through the baseline optimizer and its convergence analysis.
\paragraph{Deriving the proxy's cosh form.}
To guide the form of $\Delta L$, we use convergence analysis to describe how LR and BS affect optimization. Our baseline ladders use Muon, and \citet{pethick2025} show that an idealized form of Muon belongs to the class of methods based on a linear minimization oracle. For these methods, \citet{shulgin2026} construct an optimization proxy from bounds on a gradient-norm measure of stationarity during training, under smoothness and gradient-noise assumptions. With a fixed model, fixed momentum, constant LR, and enough training steps to neglect the transient term, they use the three-term proxy
\begin{equation}
 Q(\eta,B;D)=\frac{a_1B}{D\eta}+\frac{a_2}{\sqrt B}+a_3\eta,\qquad a_1,a_2,a_3>0.
 \label{eq:optimization_proxy}
\end{equation}
The positive coefficients absorb the fixed-model and momentum constants. The first term describes incomplete optimization within the training budget and decreases with the number of training steps $D/B$ and with LR. The second represents gradient noise and decreases with BS. The third comes from the smoothness bound and grows with the size of each update.

% Role: derive cosh from the balance between small and large LR in the proxy.
At a fixed BS, differentiating with respect to LR gives the minimum at $\eta_Q^*(B)=\sqrt{a_1B/(a_3D)}$, where the first and third terms are equal. To derive the cosh form, we express LR relative to this optimum as $r=\eta/\eta_Q^*(B)$. The sum of the first and third terms is then proportional to $r+r^{-1}$. Using $\cosh x=(e^x+e^{-x})/2$, we obtain the increase above this minimum:
\begin{equation}
 \begin{aligned}
 Q(\eta,B;D)-\min_\eta Q(\eta,B;D)
 &=\sqrt{\frac{a_1a_3B}{D}}\,(r+r^{-1}-2)\\
 &=2\sqrt{\frac{a_1a_3B}{D}}\left[\cosh\!\left(\ln\frac{\eta}{\eta_Q^*(B)}\right)-1\right].
 \end{aligned}
 \label{eq:proxy_cosh_identity}
\end{equation}
Thus, cosh expresses the balance between the inverse-LR and proportional-LR terms in log coordinates. Multiplying or dividing the optimal LR by the same factor gives the same increase in the proxy.

% Role: derive the BS dependence after optimizing LR.
Substituting the optimal LR at each BS leaves
\begin{equation}
 \min_\eta Q(\eta,B;D)=2\sqrt{\frac{a_1a_3}{D}}\sqrt B+\frac{a_2}{\sqrt B}.
 \label{eq:proxy_bs_profile}
\end{equation}
The first term increases with BS, while the second decreases. Minimizing their sum gives the optimal BS $B_Q^*=a_2\sqrt D/(2\sqrt{a_1a_3})$ and the corresponding optimal LR $\eta_Q^*=\eta_Q^*(B_Q^*)\propto D^{-1/4}$. This order of minimization follows \citet{shulgin2026} and the general-exponent treatment of \citet{su2026scaling}. With $h_Q=a_3\eta_Q^*$ and $Q^*=Q(\eta_Q^*,B_Q^*;D)=4h_Q$, we then rewrite the full proxy as
\begin{equation}
 \begin{aligned}
 Q-Q^*={}&4h_Q\left[\cosh\!\left(\frac12\ln\frac{B}{B_Q^*}\right)-1\right]\\
 &+2h_Q\sqrt{\frac{B}{B_Q^*}}\left[\cosh\!\left(\ln\frac{\eta}{\eta_Q^*(B)}\right)-1\right].
 \end{aligned}
 \label{eq:proxy_conditional_decomposition}
\end{equation}
The first term is the increase after optimizing LR at the chosen BS; the second is the additional increase from using another LR at that BS. Equations~\ref{eq:proxy_cosh_identity} and~\ref{eq:proxy_conditional_decomposition} are exact identities for the three-term proxy in Equation~\ref{eq:optimization_proxy}.

% Role: expose the local approximation that connects the exact proxy to the loss model.
The proxy's second-order expansion gives a local quadratic form for the LR--BS dependence. Writing $\widetilde{\eta}_Q=\ln(\eta/\eta_Q^*)$ and $\widetilde{B}_Q=\ln(B/B_Q^*)$, we obtain
\begin{equation}
 Q-Q^*=h_Q\left[\left(\widetilde{\eta}_Q-\frac12\widetilde{B}_Q\right)^2+\frac12\widetilde{B}_Q^2\right]
 +O\!\left((|\widetilde{\eta}_Q|+|\widetilde{B}_Q|)^3\right).
 \label{eq:proxy_quadratic}
\end{equation}
Thus the slope of optimal log LR in log BS is $1/2$, and the curvature in log BS after optimizing LR is half the curvature in log LR at the optimum. The expansion generally has third-order terms because the LR amplitude in Equation~\ref{eq:proxy_conditional_decomposition} varies with BS.

% Role: identify the transfer to validation loss and the fitted scale dependence as assumptions.
\paragraph{Adapting the proxy to validation loss.}
Using the proxy's structure to model final validation loss is a heuristic step. We use the peak LR of the WSD runs in place of the proxy's constant LR; these runs use approximate Muon updates and auxiliary AdamW (Appendix~\ref{app:optimizer}). We fit the validation-loss optima and curvature scale with the power laws in Equation~\ref{eq:power_laws}. The power laws for the optima have precedents in Section~\ref{sec:related}; the law for $h$ is our modeling extension of the fixed-model relation $h_Q=a_3\eta_Q^*\propto D^{-1/4}$. All exponents in $M$ and $D$ are fitted to validation losses.

% Role: explain the local quadratic precedent and the heuristic extension away from the optimum.
For the local shape, we retain the proxy's slope $\gamma_{\eta,B}=1/2$ and fit the curvature ratio $\rho>0$, shared across model sizes and token counts within each baseline ladder. A quadratic dependence of validation loss on log LR at fixed BS also has a fitting precedent in \citet{bjorck2025}. To describe loss farther from the predicted optima, we replace each squared log deviation $x^2$ by $2(\cosh x-1)$. The expansion $2(\cosh x-1)=x^2+x^4/12+O(x^6)$ shows that, at the same coefficient values, this choice preserves the quadratic terms while specifying faster growth away from the optimum. This replacement in both directions is the heuristic choice that gives Equation~\ref{eq:delta_l}.

% Role: identify exactly which parts of the proxy shape the adopted model changes.
Relative to the exact proxy, the adopted form makes two changes to the BS dependence. It uses an LR amplitude $h$ shared across BS values at fixed $M,D$, replacing the proxy's amplitude $h_Q\sqrt{B/B_Q^*}$. It also uses $2h\rho(\cosh\widetilde{B}-1)$ for the BS dependence after optimizing LR, replacing $4h_Q[\cosh(\widetilde{B}_Q/2)-1]$. With $\rho=1/2$ and matching optima and curvature scale, the two forms agree through second order; their higher-order terms differ. Appendix~\ref{app:lrbs_form} evaluates these modeling choices by comparing the complete forms, including a fitted form that retains the proxy shape.

% Role: derive the conditional optimum and make its exponent explicit.
\paragraph{Deriving the optimal LR at a specified BS.}
For the adopted loss model at fixed $M,D,B$, the derivative $\partial\Delta L/\partial\widetilde{\eta}=2h\sinh(\widetilde{\eta}-\gamma_{\eta,B}\widetilde{B})$ vanishes at $\widetilde{\eta}=\gamma_{\eta,B}\widetilde{B}$. The predicted optimal LR at that BS is therefore
\begin{equation}
 \begin{aligned}
 \eta^*(M,D\mid B)
 &\equiv\arg\min_\eta\widehat L(M,D,\eta,B)\\
 &=\eta^*(M,D)\left[\frac{B}{B^*(M,D)}\right]^{\gamma_{\eta,B}}.
 \end{aligned}
 \label{eq:conditional_lr}
\end{equation}
Thus $\gamma_{\eta,B}$ is both the power-law exponent of optimal LR in BS and the slope $\partial\ln\eta^*(M,D\mid B)/\partial\ln B$. It determines how the best LR changes when a different BS is chosen.

% Role: separate the lowest loss at a given BS from the increase due to LR.
Using $\widetilde{\eta}-\gamma_{\eta,B}\widetilde{B}=\ln[\eta/\eta^*(M,D\mid B)]$, the full scaling law becomes
\begin{equation}
 \begin{aligned}
 \widehat L(M,D,\eta,B)={}&L^*(M,D)+2h\rho(\cosh\widetilde{B}-1)\\
 &+2h\left[\cosh\!\left(\ln\frac{\eta}{\eta^*(M,D\mid B)}\right)-1\right].
 \end{aligned}
 \label{eq:conditional_loss}
\end{equation}
The first line is the minimum predicted loss at the chosen BS, attained by optimizing LR. The second line is the additional loss from choosing a different LR at that BS. Consequently, $\rho$ controls how quickly the minimum predicted loss rises as BS moves away from $B^*$ after LR has been adjusted. If LR is held fixed while BS changes, the optimal LR also shifts, and both terms contribute. The coefficients of the LR--BS term are fitted together in the second stage (Appendix~\ref{app:two_stage}). After fitting, Equation~\ref{eq:conditional_lr} gives the optimal LR for each specified BS, as in the fixed-BS analysis of OpenEuroLLM~\citep{openeuro2026}.

% Role: relate the local quadratic shape to the coefficients and Hessian.
\paragraph{Curvature and the role of $\rho$.}
Expanding Equation~\ref{eq:delta_l} around the predicted optima gives
\begin{equation}
 \Delta L=h\left[(\widetilde{\eta}-\gamma_{\eta,B}\widetilde{B})^2+\rho\widetilde{B}^2\right]
 +O\!\left((|\widetilde{\eta}|+|\widetilde{B}|)^4\right).
 \label{eq:lrbs_quadratic}
\end{equation}
For $h,\rho>0$, this quadratic is positive definite and has elliptical contours around the predicted optimum. The Hessian $\mathbf H$ is the matrix of second derivatives with respect to $(\widetilde{\eta},\widetilde{B})$, evaluated at this optimum with $M,D$ fixed:
\begin{equation}
 \mathbf H=2h
 \begin{pmatrix}
  1 & -\gamma_{\eta,B}\\
  -\gamma_{\eta,B} & \gamma_{\eta,B}^2+\rho
 \end{pmatrix}.
 \label{eq:lrbs_hessian}
\end{equation}
The curvature scale $h$ equals $H_{\eta\eta}/2$, half the curvature with respect to LR in log space. The off-diagonal entry describes the LR--BS coupling, with $\gamma_{\eta,B}=-H_{\eta B}/H_{\eta\eta}$. Optimizing LR at each BS leaves $\min_{\widetilde{\eta}}\Delta L=2h\rho(\cosh\widetilde{B}-1)$, whose curvature at $B^*$ is $2h\rho$. Hence
\begin{equation}
 \rho=
 \frac{\left.\dfrac{d^2}{d\widetilde{B}^2}\min_{\widetilde{\eta}}\Delta L\right|_{\widetilde{B}=0}}{H_{\eta\eta}}
 =\frac{H_{BB}-H_{\eta B}^2/H_{\eta\eta}}{H_{\eta\eta}}.
 \label{eq:lrbs_curvature_ratio}
\end{equation}
The numerator is the curvature in log BS after optimizing LR; the denominator is the curvature in log LR at the optimum. The numerator is the Schur complement of $H_{\eta\eta}$~\citep{boyd2004}.

% Role: distinguish changing BS with LR retuned from changing BS with LR fixed.
This ratio has a direct interpretation when comparing BS values near $B^*$. If LR is optimized at every BS, the leading loss increase is $h\rho\widetilde{B}^2$. If LR is instead held at $\eta^*$, the leading increase is $h(\gamma_{\eta,B}^2+\rho)\widetilde{B}^2$. The extra $\gamma_{\eta,B}^2$ arises because the fixed LR moves away from the optimal LR as BS changes. Fitting $\rho$ therefore allows the remaining sensitivity to BS after optimizing LR to be estimated separately from the fixed slope $\gamma_{\eta,B}$.

\subsection{Fitting procedure}
\label{app:two_stage}
% Role: stage 1, base surface on group minima.
The coefficients are fitted in two stages, one per term of the law. Let $g$ index the model--budget groups, $L_{g,\min}$ denote the lowest measured loss of group $g$, and $(M_g,D_g)$ denote its model scale and token count. The first stage fits the base surface of Equation~\ref{eq:coupled} to these group minima by optimizing $\phi=(L_\infty,\allowbreak\alpha_M,\allowbreak\gamma_M,\allowbreak\alpha_D,\allowbreak\gamma_D,\allowbreak\gamma_{MD})$:
\begin{equation}
 \min_\phi\sum_g
 H_\delta\!\left(\ln L^*(M_g,D_g;\phi)-\ln L_{g,\min}\right),\qquad
 H_\delta(r)=\begin{cases}r^2/2,&|r|\leq\delta,\\\delta\left(|r|-\delta/2\right),&|r|>\delta,\end{cases}
 \label{eq:base_objective}
\end{equation}
with equal weight for every group. We use the threshold of OpenEuroLLM~\citep{openeuro2026}, $\delta=0.001$. Table~\ref{tab:design_choices} compares Huber and squared losses on log-loss residuals.

% Role: stage 2, LR--BS term with the base surface fixed.
The second stage holds $L^*$ fixed and fits the parameters $\theta$ of $\Delta L$ using the measured loss of every run $i$ in every group:
\begin{equation}
 \min_\theta\sum_i
 H_\delta\!\left(\ln\!\left[L^*(M_i,D_i)+\Delta L(M_i,D_i,\eta_i,B_i;\theta)\right]-\ln L_i\right),
 \label{eq:response_objective}
\end{equation}
with the same Huber loss and threshold as Equation~\ref{eq:base_objective} and equal weight for every run. Both fitting stages use bounded nonlinear optimization in log-normalized coordinates, retaining the solution with the lowest objective across several random starts. Tables~\ref{tab:coupled_parameters} and~\ref{tab:response_parameters} in Section~\ref{sec:analysis} report the fitted base-surface coefficients and LR--BS parameters.

\subsection{Additional fit diagnostics}
\label{app:fit_quality}
\label{app:error_metrics}
\label{app:lrbs_maps}
% Role: which runs the diagnostics evaluate and which run checks are separate.
We compare predicted and measured losses for the swept runs used for fitting. The extrapolation runs and Phase~II pretraining runs are excluded from fitting; their loss predictions are checked separately in Section~\ref{sec:fit_quality} and Appendix~\ref{app:pretraining_sweep}.

% Role: error-metric definitions.
For $n$ runs with relative errors $e_i=\widehat L_i/L_i-1$, the all-point MAPE is $(1/n)\sum_i|e_i|$. The best-point MAPE applies the same formula to the lowest-loss run of each model--budget group, and P95 is the 95th percentile of $|e_i|$.

% Role: OLMo3 LR--BS response map.
For the OLMo3 ladder, Figure~\ref{fig:response_maps_olmo} compares the fitted and measured dependence of loss on LR and BS, complementing the Qwen3.5 results in Figure~\ref{fig:response_maps}. The predictions use the coefficients in Table~\ref{tab:response_parameters}.

\input{figures/response_maps_olmo_figure.tex}

\subsection{Compute-optimal model size, training data, and predicted optima}
\label{app:allocation}
\label{app:predicted_optima}
% Role: derive the optimal model size and training token count from the fitted base surface.
Following the Chinchilla analysis~\citep{hoffmann2022} and its extension in Skaling~\citep{skaling2026}, we minimize $L^*(M,D)$ subject to $MD=C$ to obtain the optimal model size and training token count. Substituting $D=C/M$ into Equation~\ref{eq:coupled} gives the stationarity condition $\gamma_M\alpha_M (M^*)^{-\gamma_M}=\gamma_D\alpha_D (D^*)^{-\gamma_D}$. The positive outer exponent does not change the minimizer, so with $M^*D^*=C$,
\begin{align}
 M^*(C)&=\left(\frac{\gamma_M\alpha_M}{\gamma_D\alpha_D}\right)^{1/(\gamma_M+\gamma_D)}
 C^{\gamma_D/(\gamma_M+\gamma_D)},&D^*(C)&=C/M^*(C),\label{eq:allocation_exact}\\
 D^*(M)&=\left(\frac{\gamma_D\alpha_D}{\gamma_M\alpha_M}\right)^{1/\gamma_D}
 M^{\gamma_M/\gamma_D}.\label{eq:model_optimal_tokens}
\end{align}
Equation~\ref{eq:allocation_exact} gives the optimal model scale and token count for a given budget, while Equation~\ref{eq:model_optimal_tokens} gives the token count on the compute-optimal curve at a given model scale. Evaluating Equation~\ref{eq:power_laws} at the optimal model size and training token count then gives the predicted optimal LR and BS. If the BS is fixed, minimizing over the LR leaves the term $2\rho h\,[\cosh\ln(B/B^*)-1]$, which must also be included when re-optimizing model size and training token count.

% Role: derive the compute--loss form along the compute-optimal curve.
Substituting Equation~\ref{eq:allocation_exact} into Equation~\ref{eq:coupled} makes both the model term and the data term proportional to $C^{-\gamma_M\gamma_D/(\gamma_M+\gamma_D)}$. The resulting loss has the form of Equation~\ref{eq:compute_law}, with $\gamma_C=\gamma_{MD}\gamma_M\gamma_D/(\gamma_M+\gamma_D)$.

% Role: realizable model sizes and training token counts for Setting~1.
Setting~1 of Section~\ref{sec:phase1} uses the baseline's compute-optimal curve to choose the model size and training token count. A realizable shape of the proposed architecture has model scale $M$, which rarely satisfies $M=M^*(C_j)$ exactly for a compute budget $C_j$. The training token count is then set in one of two ways:
\begin{align}
 &\text{size first:} && D=D^*(M), && \text{training budget } M\,D^*(M);\label{eq:size_first_allocation}\\
 &\text{budget first:} && M\approx M^*(C_j), && D=C_j/M,\ \text{training budget } C_j.\label{eq:budget_first_allocation}
\end{align}
In the size-first construction, the model scale is fixed and the token count should be chosen so that the model scale and token count lie exactly on the baseline's optimal curve; the training budget can differ from $C_j$. In the budget-first construction, a model scale near the optimum should be chosen, and the token count should be set to keep the budget at $C_j$. The resulting model size and training token count differ slightly from their optimal values.

% Role: predicted optima at the compute budgets and beyond.
Table~\ref{tab:optima} gives the compute-optimal model sizes and training token counts from the fitted base surfaces and the predicted optimal LR and BS of the adopted fixed-slope laws at the five compute budgets and at $1\times10^{21}$ FLOPs. These continuous predictions complement the best measured recipes of Appendix~\ref{app:recipes}. The predicted model sizes and training token counts should be used to choose model shapes in Setting~1 and the ranges of model sizes searched in Setting~2 and Setting~3. The predicted optima provide the initial LR and BS for the Setting~3 search.
\input{tables/fitted_optima.tex}

\subsection{Extrapolation recipes}
\label{app:extrapolation_recipes}
% Role: recipes of the extrapolation runs and what those runs test.
We test the frozen fixed-slope laws of Section~\ref{sec:analysis} by comparing their predictions with the measured losses of the extrapolation runs (Table~\ref{tab:high_compute}). Table~\ref{tab:extrapolation_recipes} gives the LR and BS used in these runs. A single recipe is used for each model--budget pair, and the predicted LR--BS term ranges from 0.00004 to 0.0018. The base surface $L^*$ alone predicts loss with a MAPE of 0.33\% on these runs, so they mainly test the base surface. For the largest swept model of each baseline ladder, we also predict the loss reduction from its best measured $1\times10^{20}$ recipe to its extrapolation recipe with a relative error below 6\%.
\input{tables/extrapolation_recipes.tex}

%% file: figures/response_maps_olmo_figure.tex
\begin{figure}[tp]
\centering
\includegraphics[width=\linewidth]{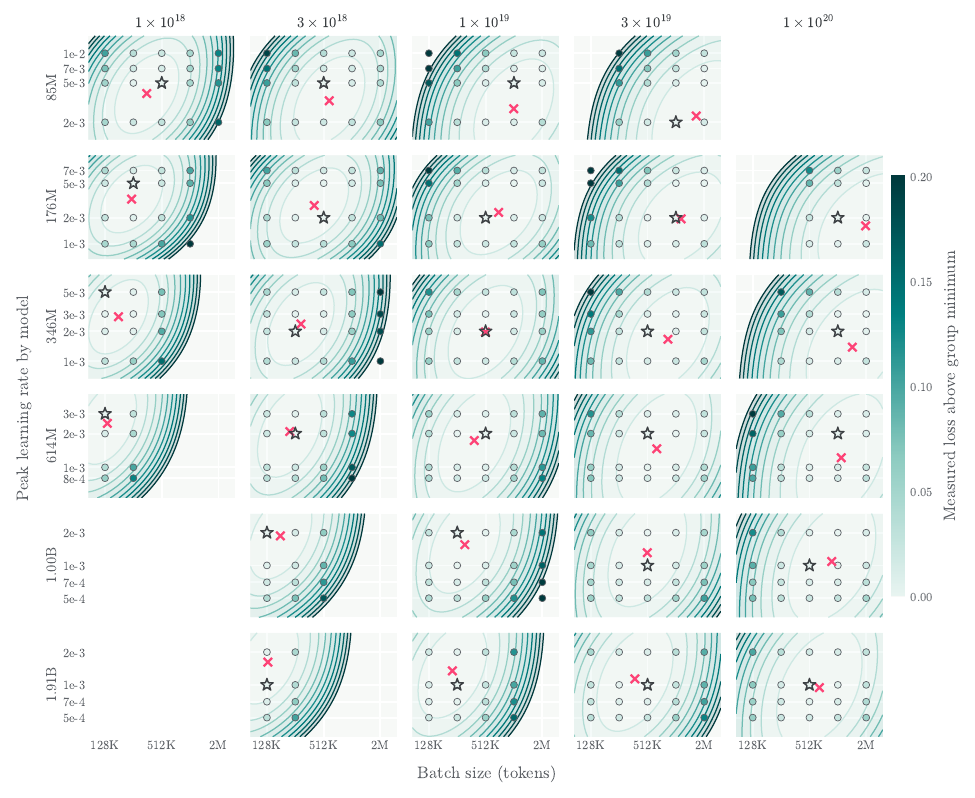}
\caption{LR--BS dependence of loss in the OLMo3 baseline ladder. Colors show the measured loss above each group minimum, and contours show the fitted LR--BS term $\Delta L$ at intervals of 0.02. Crosses mark predicted optima and stars the best measured configurations; empty panels are outside the sweep. Table~\ref{tab:response_parameters} gives the fitted laws.}
\label{fig:response_maps_olmo}
\end{figure}

%% file: tables/fitted_optima.tex
\begin{table}[tp]
\caption{Compute-optimal model sizes and training token counts, together with predicted optima of the adopted fixed-slope scaling laws. Daggers mark model scales beyond the swept range. The predictions at $1\times10^{21}$ FLOPs are extrapolations beyond the swept budgets.}
\label{tab:optima}
\centering\footnotesize
\input{tables/generated/fitted_optima.tex}
\end{table}

%% file: tables/generated/fitted_optima.tex
% Generated by scripts/make_baseline_figures.py from frozen evidence; do not edit by hand.
\begin{tabular}{llrrrrr}
\toprule
Baseline & FLOPs & $M^*$ (GFLOPs/token) & $D^*$ & $\eta^*$ & $B^*$ & $L^*$\\
\midrule
Qwen3.5 & $1\times10^{18}$ & 1.02 & 978M & 3.82e-3 & 315K & 3.4649\\
Qwen3.5 & $3\times10^{18}$ & 1.70 & 1.76B & 2.84e-3 & 409K & 3.2457\\
Qwen3.5 & $1\times10^{19}$ & 2.99 & 3.35B & 2.05e-3 & 545K & 3.0370\\
Qwen3.5 & $3\times10^{19}$ & 4.98 & 6.02B & 1.52e-3 & 707K & 2.8716\\
Qwen3.5 & $1\times10^{20}$ & 8.73 & 11.5B & 1.10e-3 & 942K & 2.7142\\
Qwen3.5 & $1\times10^{21}$ & 25.5$^\dagger$ & 39.2B & 5.87e-4 & 1.59M & 2.4706\\
\addlinespace
OLMo3 & $1\times10^{18}$ & 0.981 & 1.02B & 3.53e-3 & 286K & 3.5147\\
OLMo3 & $3\times10^{18}$ & 1.61 & 1.86B & 2.65e-3 & 371K & 3.2949\\
OLMo3 & $1\times10^{19}$ & 2.78 & 3.60B & 1.94e-3 & 491K & 3.0812\\
OLMo3 & $3\times10^{19}$ & 4.56 & 6.58B & 1.46e-3 & 636K & 2.9084\\
OLMo3 & $1\times10^{20}$ & 7.85 & 12.7B & 1.07e-3 & 843K & 2.7405\\
OLMo3 & $1\times10^{21}$ & 22.2$^\dagger$ & 45.0B & 5.85e-4 & 1.41M & 2.4726\\
\bottomrule
\end{tabular}

%% file: tables/extrapolation_recipes.tex
\begin{table}[tp]
\caption{LR and BS used for the extrapolation runs whose measured and predicted losses are compared in Table~\ref{tab:high_compute}.}
\label{tab:extrapolation_recipes}
\centering\small
\input{tables/generated/extrapolation_recipes.tex}
\end{table}

%% file: tables/generated/extrapolation_recipes.tex
% Generated by scripts/make_baseline_figures.py from frozen evidence; do not edit by hand.
\begin{tabular}{llrr}
\toprule
Baseline & Model & $\eta$ & $B$\\
\midrule
Qwen3.5 & 1.79B & 8e-4 & 1M\\
Qwen3.5 & 2.64B & 6e-4 & 1M\\
OLMo3 & 1.91B & 8e-4 & 1M\\
OLMo3 & 3.36B & 6e-4 & 2M\\
\bottomrule
\end{tabular}

%% file: sections/13_appendix_formula_ablations.tex
\section{Formula Design Choices and Ablations}
\label{app:ablations}
% Role: purpose and shared setup of the ablations.
The ablations examine the base surface, fitting procedure, LR--BS term, and the estimation of the optima to explain the scaling-law choices summarized in Table~\ref{tab:design_choices}. Within each baseline ladder, alternative base surfaces are fitted to the same group minima, and alternative LR--BS forms are fitted to losses from the same swept runs. Comparisons use the MAPE defined in Appendix~\ref{app:error_metrics}.

\input{tables/design_choices.tex}

\subsection{Coupled versus additive base surface}
\label{app:additive}
% Role: why the coupled form, and how much the coupling matters.
The coupled surface of Equation~\ref{eq:coupled} reduces to the additive form of Chinchilla~\citep{hoffmann2022} when $\gamma_{MD}=1$. Skaling motivates the coupling through the mixed derivative~\citep{skaling2026}. With $X=\alpha_MM^{-\gamma_M}$ and $Y=\alpha_DD^{-\gamma_D}$,
\begin{equation}
 \frac{\partial^2L^*}{\partial\ln M\,\partial\ln D}
 =\gamma_{MD}(\gamma_{MD}-1)\gamma_M\gamma_D\,XY(X+Y)^{\gamma_{MD}-2},
 \label{eq:base_mixed_derivative}
\end{equation}
which vanishes for the additive form and is negative for $0<\gamma_{MD}<1$. The coupled form thus allows the effect of model scale to depend on training data. To test the value of this interaction, we fit the additive form to the same group minima, retaining the same LR--BS term.

% Role: fitted coefficients and fit errors of the two surfaces.
Table~\ref{tab:additive_parameters} gives the coefficients and best-point MAPE of both base surfaces. Figure~\ref{fig:isoflops} compares their IsoFLOP curves with the measured minima, and Section~\ref{sec:fit_quality} discusses the comparison.
\input{tables/additive_parameters.tex}

\subsection{Fitting order}
\label{app:fit_order}
% Role: setup of the fitting-order ablation.
In the two-stage procedure, we fit $L^*$ first and hold it fixed while fitting $\Delta L$ (Appendix~\ref{app:two_stage}). We test whether fitting the two terms simultaneously gives more accurate loss predictions. Both approaches use the same functional forms and equal weighting. The LR--BS term takes the log-quadratic form $h[(\widetilde{\eta}-\gamma_{\eta,B}\widetilde{B})^2+\rho\widetilde{B}^2]$ (Appendix~\ref{app:lrbs_term}); its slope $\gamma_{\eta,B}$ and curvature ratio $\rho$ are fitted in both approaches. Table~\ref{tab:fit_order} reports errors on the swept runs and on runs excluded from fitting.

% Role: fitting-order trade-off and its relevance to optimal model size and training token count.
On the swept runs, the base surface from two-stage fitting predicts the group minima more accurately, while simultaneous fitting gives similar or lower loss error across all points by allowing $L^*$ and $\Delta L$ to adjust together. For runs excluded from fitting, simultaneous fitting predicts loss more accurately on the extrapolation runs but less accurately on the 150B-token pretraining sweep. We retain two-stage fitting because it preserves the fit of the base surface to the group minima used to determine the IsoFLOP curves and compute-optimal model sizes and training token counts.
\input{tables/fit_order.tex}

\subsection{Form of the LR--BS term}
\label{app:lrbs_form}
% Role: forms compared and how they are scored.
We compare four LR--BS forms to test how accurately they predict loss and locate the optimal LR and BS (Table~\ref{tab:lrbs_forms}). The alternatives are a form that retains the proxy shape (Appendix~\ref{app:lrbs_term}), the log-quadratic form, and cosh forms with fitted or fixed LR--BS slopes. The log-quadratic form generalizes the proxy's second-order expansion in Equation~\ref{eq:proxy_quadratic} by fitting the slope and curvature ratio. Each alternative uses power laws for its optima and amplitude, with optimal BS depending on both $M$ and $D$ (Section~\ref{sec:literature}). We fit these forms using the same fitted base surface and losses from all swept runs. Alongside loss MAPE, the table reports the MAPE of the predicted optima relative to the best measured configuration of each model--budget group.

% Role: form comparison results and the adopted choice.
The form that retains the proxy shape has the largest loss error across all swept runs. Compared with the log-quadratic form, both cosh forms reduce all-point loss MAPE from 0.28\% to 0.25\% for Qwen3.5 and from 0.31\% to 0.27\% for OLMo3. Fitting the slope gives $\gamma_{\eta,B}=0.77$ for Qwen3.5 and 0.51 for OLMo3, with nearly the same loss MAPE as fixing it. For Qwen3.5, however, fitting the slope raises the MAPE of the predicted optimal LR to 50.0\%, compared with 34.5\% for the log-quadratic form. Compared with fitting the cosh slope, fixing it gives an LR MAPE of 35.1\% with one fewer parameter and nearly the same loss MAPE. We therefore adopt the cosh form with $\gamma_{\eta,B}=1/2$ for both baseline ladders. Section~\ref{sec:fit_quality} compares the resulting contours with the measured LR--BS dependence.

\input{tables/lrbs_forms.tex}

\subsection{Shared versus per-group estimation of the optima}
\label{app:per_group}
% Role: comparison with the estimation route of OpenEuroLLM.
We test whether estimating the optima separately for each model--budget group is more accurate than using the shared LR--BS term. OpenEuroLLM estimates the optima by fitting a quadratic surface to the losses in each group~\citep{openeuro2026}. We fit each quadratic in log LR and log BS, retaining fits with a full-rank design matrix, a positive-definite quadratic, and a minimum strictly inside the convex hull of the tested points. These fits cover 24 of the 27 groups for Qwen3.5 and 25 of 27 for OLMo3. We then fit power laws in $M$ and $D$ to the retained minima.

% Role: errors of the two approaches against the per-group minima.
When evaluated against these minima, the per-group power laws attain lower MAPE than the shared LR--BS term, which is fitted to losses from all swept runs. The LR errors are 11.3\% versus 16.6\% for Qwen3.5 and 5.4\% versus 12.5\% for OLMo3. The BS errors are 9.9\% versus 10.5\% for Qwen3.5 and 6.2\% versus 7.5\% for OLMo3.

% Role: why the shared LR--BS term remains the adopted description.
The per-group power laws, however, predict only the optimal LR and BS. The shared LR--BS term also predicts loss away from its optima and the optimal LR at a fixed BS (Equation~\ref{eq:conditional_lr}). We therefore retain the shared term, using the per-group minima as a separate check of its predicted optima. At an imposed BS, the LR predicted by the shared term should be used to initialize the Setting~3 search (Section~\ref{sec:phase1}).

%% file: tables/design_choices.tex
\begin{table}[tp]
\caption{Design choices of the scaling law for the baseline ladders.}
\label{tab:design_choices}
\centering\small
\begin{tabularx}{\linewidth}{>{\raggedright\arraybackslash}p{2.4cm}YY}
\toprule
Component & Alternatives & Selected\\
\midrule
Base surface & Additive form of Chinchilla~\citep{hoffmann2022} ($\gamma_{MD}=1$); coupled form of Skaling and OpenEuroLLM~\citep{skaling2026,openeuro2026} & Coupled form, Equation~\ref{eq:coupled} (Appendix~\ref{app:additive})\\
Fitting objective & Squared log-loss residuals; Huber loss on log-loss residuals, as in Skaling and OpenEuroLLM & Huber loss with $\delta=0.001$ when fitting $L^*$ and $\Delta L$, Equations~\ref{eq:base_objective} and~\ref{eq:response_objective}; best-point MAPE of $L^*$ is 0.13\% (Qwen3.5) and 0.11\% (OLMo3), compared with 0.13\% and 0.12\% for the squared loss\\
Fitting order & $L^*$ and $\Delta L$ fitted together; $L^*$ fitted first and held fixed while $\Delta L$ is fitted & Two-stage fit (Appendix~\ref{app:fit_order})\\
Estimating the optima & Quadratic surfaces fitted to each group's losses, then power laws fitted to the minima of those surfaces, following OpenEuroLLM; one LR--BS term fitted to losses from all runs & One shared LR--BS term (Appendix~\ref{app:per_group})\\
Form of $\Delta L$ & Proxy shape of \citeauthor{shulgin2026}~\citep{shulgin2026} with fitted power laws for its optima and amplitude; log-quadratic form based on the proxy's second-order expansion, with fitted slope and curvature ratio; cosh form with a fitted slope; cosh form with the proxy's slope $\gamma_{\eta,B}=1/2$ & Cosh form with $\gamma_{\eta,B}=1/2$, Equation~\ref{eq:delta_l} (Appendix~\ref{app:lrbs_form})\\
Optimal BS & $B^*$ depending on $D$ only, as in StepLaw~\citep{steplaw2025}; $B^*$ depending on $M$ and $D$, as in OpenEuroLLM & Depends on $M$ and $D$ for all baseline ladders\\
\bottomrule
\end{tabularx}
\end{table}

%% file: tables/additive_parameters.tex
\begin{table}[tp]
\caption{Coupled and additive base surfaces fitted to the same group minima. MAPE is evaluated on those minima; $M$ is in FLOPs per token and $D$ in tokens. Figure~\ref{fig:additive_curves} compares their IsoFLOP curves with measurements.}
\label{tab:additive_parameters}
\centering\footnotesize
\input{tables/generated/additive_parameters.tex}
\end{table}

%% file: tables/generated/additive_parameters.tex
% Generated by scripts/make_baseline_figures.py from frozen evidence; do not edit by hand.
\begin{tabular}{llrrrrrrr}
\toprule
Baseline & Form & $L_\infty$ & $\alpha_M$ & $\gamma_M$ & $\alpha_D$ & $\gamma_D$ & $\gamma_{MD}$ & MAPE\\
\midrule
Qwen3.5 & Coupled & 1.722 & $6.77\times10^{4}$ & 0.520 & $1.90\times10^{4}$ & 0.453 & 0.505 & 0.13\%\\
Qwen3.5 & Additive & 2.300 & $3.61\times10^{4}$ & 0.537 & $1.28\times10^{3}$ & 0.364 & 1.000 & 0.48\%\\
\addlinespace
OLMo3 & Coupled & 1.488 & $3.96\times10^{5}$ & 0.576 & $5.99\times10^{4}$ & 0.475 & 0.402 & 0.11\%\\
OLMo3 & Additive & 2.348 & $8.52\times10^{4}$ & 0.580 & $1.37\times10^{3}$ & 0.365 & 1.000 & 0.62\%\\
\bottomrule
\end{tabular}

%% file: tables/fit_order.tex
\begin{table}[tp]
\caption{Fitting-order ablation with a log-quadratic LR--BS term whose slope and curvature ratio are fitted. Two-stage fitting estimates $L^*$ from group minima and holds it fixed while fitting $\Delta L$; simultaneous fitting estimates both from the losses of all swept runs.}
\label{tab:fit_order}
\centering\footnotesize
\input{tables/generated/fit_order.tex}
\end{table}

%% file: tables/generated/fit_order.tex
% Generated by scripts/make_baseline_figures.py from frozen evidence; do not edit by hand.
\begin{tabular}{lrrrrrr}
\toprule
& \multicolumn{6}{c}{MAPE}\\
& \multicolumn{2}{c}{$L^*$, best-point} & \multicolumn{2}{c}{$\widehat L$, best-point} & \multicolumn{2}{c}{$\widehat L$, all-point}\\
\cmidrule(lr){2-3}\cmidrule(lr){4-5}\cmidrule(lr){6-7}Baseline & Two-stage & Simultaneous & Two-stage & Simultaneous & Two-stage & Simultaneous\\
\midrule
\multicolumn{7}{l}{\textit{Swept runs}}\\
Qwen3.5 & 0.13\% & 0.16\% & 0.21\% & 0.21\% & 0.28\% & 0.27\%\\
OLMo3 & 0.11\% & 0.24\% & 0.19\% & 0.18\% & 0.31\% & 0.29\%\\
\midrule
\multicolumn{7}{l}{\textit{Runs excluded from fitting}}\\
\multicolumn{5}{l}{Qwen3.5 extrapolation runs} & 0.27\% & 0.23\%\\
\multicolumn{5}{l}{OLMo3 extrapolation runs} & 0.39\% & 0.07\%\\
\multicolumn{5}{l}{Qwen3.5 150B-token pretraining} & 0.96\% & 1.17\%\\
\bottomrule
\end{tabular}

%% file: tables/lrbs_forms.tex
\begin{table}[tp]
\caption{Each alternative LR--BS form is fitted using the same base surface and losses from all swept runs. The Best-point and All-point columns report loss MAPE on group minima and all runs; the $\eta^*$ and $B^*$ columns report MAPE against the best measured configuration of each group. Parameter counts include the base surface. Bold marks the adopted form.}
\label{tab:lrbs_forms}
\centering\footnotesize
\setlength{\tabcolsep}{4.5pt}
\input{tables/generated/lrbs_forms.tex}
\end{table}

%% file: tables/generated/lrbs_forms.tex
% Generated by scripts/make_baseline_figures.py from frozen evidence; do not edit by hand.
\begin{tabular}{lrrrrrrrrr}
\toprule
& & \multicolumn{4}{c}{Qwen3.5: MAPE} & \multicolumn{4}{c}{OLMo3: MAPE}\\
\cmidrule(lr){3-6}\cmidrule(lr){7-10}Form of $\Delta L$ & Parameters & \shortstack{Best-\\point} & \shortstack{All-\\point} & $\eta^*$ & $B^*$ & \shortstack{Best-\\point} & \shortstack{All-\\point} & $\eta^*$ & $B^*$\\
\midrule
Proxy shape & 15 & 0.29\% & 0.46\% & 40.57\% & 73.85\% & 0.26\% & 0.53\% & 28.90\% & 75.58\%\\
Log-quadratic & 17 & 0.21\% & 0.28\% & 34.49\% & 23.19\% & 0.19\% & 0.31\% & 22.47\% & 30.41\%\\
Cosh, fitted slope & 17 & 0.21\% & 0.25\% & 50.04\% & 24.33\% & 0.17\% & 0.27\% & 23.19\% & 26.59\%\\
\textbf{Cosh, $\gamma_{\eta,B}=1/2$} & 16 & \textbf{0.22\%} & \textbf{0.25\%} & \textbf{35.10\%} & \textbf{21.42\%} & \textbf{0.17\%} & \textbf{0.27\%} & \textbf{23.11\%} & \textbf{26.68\%}\\
\bottomrule
\end{tabular}

%% file: sections/15_appendix_long_training.tex
\section{Phase II Baselines: Configuration and Results}
\label{app:phase2}
\subsection{Configuration and evaluation}
\label{app:phase2_config}
% Role: baseline shapes and the pretraining schedule.
The Qwen3.5 and OLMo3 Phase~II baselines use the 24-layer shapes in Table~\ref{tab:exact_shapes} and the shared optimizer and initialization settings in Appendix~\ref{app:recipes}. Pretraining uses 8K context and a 2,000-update warmup. A square-root decay over the final 20\% of updates reduces the LR to 10\% of its peak.

% Role: initialization, data, and schedule for long-context mid-training.
Each long-context mid-training run then restores the model and optimizer state of the selected pretrained checkpoint and trains on about 10B tokens of the Dolma~3 Longmino Mix (50B) at 64K context. This stage uses a 200-update warmup followed by a stable LR and a square-root decay to zero over the final 20\% of updates.

% Role: document boundaries define the long-context training examples.
Long-context mid-training retains document boundaries within the 64K sequences~\citep{olmo3}. Attention is causal within each document and masked across documents, and rotary positions restart at each boundary.

% Role: the positional extension differs between the two architectures.
At 64K context, Qwen3.5 retains its pretraining RoPE base of $1\times10^7$ and rotary fraction of 0.25 without further positional scaling. By contrast, OLMo3 applies YaRN scaling~\citep{peng2024yarn} with factor 8 to the rotary embeddings of its full-attention layers; its sliding-window layers retain their pretraining position encoding.

% Role: the selected recipe, its reuse, and evaluation of the resulting checkpoints.
The LR and BS selected in Appendices~\ref{app:pretraining_sweep} and~\ref{app:midtraining} form the Qwen3.5 baseline recipe, which OLMo3 reuses for both stages (Table~\ref{tab:baseline_recipe}). Section~\ref{sec:baseline_capabilities} describes the evaluation of the resulting checkpoints and reports the principal Base Main and Base Long-Context results.
\input{tables/baseline_recipe.tex}

\subsection{Pretraining sweep}
\label{app:pretraining_sweep}
% Role: design of the pretraining sweep and its starting point.
The Qwen3.5 pretraining sweep starts from the predicted optima in Table~\ref{tab:phase2_predictions}. At BS 4M, the tested peak LR values extend up to 5e-3 and bracket the LR of the lowest-loss recipe. Fewer LR values are tested at BS 2M and 8M. Table~\ref{tab:stage1_sweep} lists every run with its Base Main scores and the loss predicted by the frozen scaling law. Figure~\ref{fig:coverage} shows the sweep's TPP relative to the baseline ladders.

% Role: how the sweep departs from the ladder regime.
The pretraining sweep extends beyond the baseline ladder in TPP and uses a different warmup schedule. The ladder uses a fixed token count for warmup, whereas the sweep uses a fixed update count. The sweep's warmup fraction therefore increases with BS (Appendix~\ref{app:phase2_config}).

% Role: frozen-law predictions on the pretraining sweep.
The frozen scaling law overpredicts every measured loss, with a MAPE of 1.02\%. At the selected recipe, the overprediction is 1.52\% for the full law and 1.03\% for the base surface alone. The selected LR lies above the predicted optimum at the same BS, so the LR--BS term increases the overprediction (Section~\ref{sec:pretraining}).

% Role: loss versus capabilities when selecting the pretraining recipe.
Downstream task rankings differ from the ranking by final validation loss. Table~\ref{tab:stage1_downstream} compares the lowest-loss recipe with close alternatives on benchmarks evaluated separately from the Base Main suite. A higher-loss alternative performs best on each benchmark shown, and the best recipe varies by task. We consider these task-dependent rankings when selecting the pretraining recipe (Section~\ref{sec:pretraining}).
\input{tables/stage1_downstream.tex}

\subsection{Effect of the decay}
\label{app:rank_reversal}
% Role: why runs are compared only after the full schedule.
The pretraining runs change rank during LR decay (Figure~\ref{fig:stage1_trajectories}). Before decay, loss increases with peak LR at every tested BS; across all runs, 8e-4/4M ranks first and 3e-3/4M eighth. During decay, loss falls by 0.09 to 0.18, generally more at higher peak LR, consistent with the sharp loss reduction during the cooldown of WSD schedules~\citep{hagele2024}. At the final checkpoint, 3e-3/4M ranks first and 8e-4/4M last. This reversal supports comparing the final checkpoints of complete schedules, as required in Phase~II.
\input{figures/stage1_trajectories_figure.tex}

\subsection{Long-context mid-training}
\label{app:midtraining}
% Role: design of the mid-training sweep and why it searches directly.
The scaling laws are fitted to losses measured in the baseline ladders and do not apply to long-context mid-training. We therefore sweep LR and BS directly for Qwen3.5 rather than starting from predicted optima. At the pretraining BS of 4M, peak LR ranges from the final pretraining LR, 3e-4, to the pretraining peak, 3e-3; fewer LR values are tested at a doubled BS of 8M. Table~\ref{tab:midtraining_sweep} reports Base Main scores and 64K RULER and HELMET scores for every run.
\input{tables/midtraining_sweep.tex}

% Role: interaction of LR and BS in mid-training.
Doubling BS lowers the 64K RULER and HELMET scores at LR 1e-3 but raises them at 3e-3. At BS 8M, both scores increase with LR up to the largest tested value.

% Role: scores as a function of evaluation length and by category.
Long-context mid-training also preserves RULER performance as the evaluation length grows (Figure~\ref{fig:midtraining_sweep}a and Table~\ref{tab:context_lengths}). At 64K, beyond the 8K pretraining window, the pretrained checkpoint retains 27\% of its 4K RULER score, compared with 66\%--71\% for the mid-trained checkpoints. Among the tested recipes, the selected run has the highest RULER score at 64K, the length targeted by Phase~II, while 3e-3/4M scores highest at 4K and 8K. Table~\ref{tab:context_categories} gives the 64K scores by category.
\input{tables/context_lengths.tex}
\input{tables/context_categories.tex}

% Role: reuse of the Qwen3.5 recipe and the resulting OLMo3 scores.
We also reuse the Qwen3.5 baseline recipe for OLMo3 without further tuning. Table~\ref{tab:olmo3_long} reports the resulting Base Main scores and the 64K RULER and HELMET scores after each training stage.
\input{tables/olmo3_long.tex}

%% file: tables/baseline_recipe.tex
\begin{table}[tp]
\caption{Phase~II baseline recipe, selected for Qwen3.5 and reused for OLMo3 without tuning (Section~\ref{sec:phase2_baselines}).}
\label{tab:baseline_recipe}
\centering\small
\input{tables/generated/baseline_recipe.tex}
\end{table}

%% file: tables/generated/baseline_recipe.tex
% Generated by scripts/make_long_training_figures.py from frozen evidence; do not edit by hand.
\begin{tabular}{lrrrrrr}
\toprule
Stage & Context & Tokens & $\eta$ & $B$ & Warmup & Final $\eta$\\
\midrule
Pretraining & 8K & 157.4B & 3e-3 & 4M & 2,000 updates & 3e-4\\
Long-context mid-training & 64K & 10.7B & 1e-3 & 4M & 200 updates & 0\\
\bottomrule
\end{tabular}

%% file: tables/stage1_downstream.tex
\begin{table}[tp]
\caption{Qwen3.5 benchmark results at the end of pretraining for the selected recipe and alternatives from Table~\ref{tab:stage1_sweep}. MMLU uses length-normalized accuracy; bits per byte is measured on the reference answer. These evaluations are separate from the Base Main suite.}
\label{tab:stage1_downstream}
\centering\small
\input{tables/generated/stage1_downstream.tex}
\end{table}

%% file: tables/generated/stage1_downstream.tex
% Generated by scripts/make_long_training_figures.py from frozen evidence; do not edit by hand.
\begin{tabular}{lrrr}
\toprule
Metric & 3e-3/4M & 2e-3/4M & 2e-3/2M\\
\midrule
Validation loss $L$ & 2.4878 & 2.4933 & 2.4958\\
\midrule
ARC-Challenge accuracy $\uparrow$ & 24.8\% & 25.5\% & \textbf{27.2\%}\\
MMLU humanities accuracy $\uparrow$ & 25.4\% & 24.6\% & \textbf{25.9\%}\\
MMLU social sciences accuracy $\uparrow$ & 26.4\% & 26.6\% & \textbf{30.0\%}\\
MMLU STEM accuracy $\uparrow$ & 25.1\% & 27.0\% & \textbf{27.8\%}\\
Basic-skills coding accuracy $\uparrow$ & 60.7\% & \textbf{64.3\%} & 63.8\%\\
MBPP bits per byte $\downarrow$ & 0.590 & \textbf{0.588} & 0.600\\
Minerva Math bits per byte $\downarrow$ & 0.610 & \textbf{0.607} & 0.616\\
\bottomrule
\end{tabular}

%% file: figures/stage1_trajectories_figure.tex
\begin{figure}[tp]
\centering
\includegraphics[width=.78\linewidth]{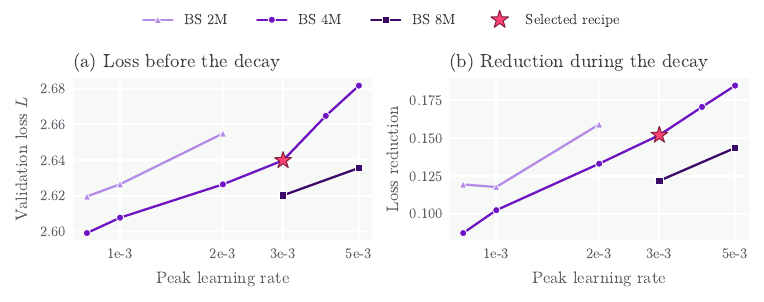}
\caption{Loss measured before decay (a) and reduction during decay (b) in the Qwen3.5 pretraining sweep. Subtracting the reduction from the loss before decay gives the final loss in Table~\ref{tab:stage1_sweep}, showing how recipe rankings change during decay.}
\label{fig:stage1_trajectories}
\end{figure}

%% file: tables/midtraining_sweep.tex
\begin{table}[tp]
\caption{Qwen3.5 long-context mid-training sweep: Base Main scores and 64K RULER and HELMET scores for every run. Long-context scores for the pretrained checkpoint are reference measurements. Base Main input limits are 8K for the pretrained checkpoint and 64K for the mid-trained checkpoints. Figure~\ref{fig:midtraining_sweep} plots the Base Main and long-context scores, and Table~\ref{tab:context_lengths} gives long-context results at each evaluation length.}
\label{tab:midtraining_sweep}
\centering\small
\input{tables/generated/midtraining_sweep.tex}
\end{table}

%% file: tables/generated/midtraining_sweep.tex
% Generated by scripts/make_long_training_figures.py from frozen evidence; do not edit by hand.
\begin{tabular}{llrrrrrrr}
\toprule
& & \multicolumn{5}{c}{Base Main} & \multicolumn{2}{c}{64K}\\
\cmidrule(lr){3-7}\cmidrule(lr){8-9}$\eta$ & $B$ & Score & Math & MC-STEM & MC-NonSTEM & GenQA & RULER & HELMET\\
\midrule
\multicolumn{2}{l}{Pretrained checkpoint} & 27.9 & 3.2 & 27.0 & 28.8 & 52.6 & 17.5 & 23.3\\
\midrule
3e-4 & 4M & 44.5 & 15.5 & 49.8 & 56.4 & 56.2 & 51.9 & 37.8\\
5e-4 & 4M & 45.5 & 17.3 & 50.6 & 57.9 & 56.3 & 52.6 & 38.4\\
1e-3 & 4M & 46.3 & 18.6 & 51.3 & 59.3 & 56.0 & \textbf{57.0} & 39.1\\
1.5e-3 & 4M & 46.3 & 19.8 & 50.4 & 59.2 & 55.8 & 56.0 & \textbf{40.0}\\
3e-3 & 4M & 45.5 & \textbf{20.7} & 49.3 & 58.0 & 54.2 & 53.7 & 37.5\\
\midrule
1e-3 & 8M & 45.9 & 17.7 & \textbf{51.4} & 58.3 & \textbf{56.4} & 53.8 & 37.2\\
1.5e-3 & 8M & \textbf{46.4} & 19.6 & 50.7 & \textbf{59.5} & 55.7 & 54.7 & 38.3\\
3e-3 & 8M & 45.7 & 20.1 & 50.0 & 57.9 & 54.6 & 56.4 & \textbf{40.0}\\
\bottomrule
\end{tabular}

%% file: tables/context_lengths.tex
\begin{table}[tp]
\caption{Long-context scores for the Qwen3.5 mid-training sweep and the OLMo3 baseline checkpoints; scores for the pretrained checkpoints are reference measurements. Figure~\ref{fig:midtraining_sweep}a,b plots the sweep results; Table~\ref{tab:context_categories} breaks down the 64K scores by category.}
\label{tab:context_lengths}
\centering\footnotesize
\input{tables/generated/context_lengths.tex}
\end{table}

%% file: tables/generated/context_lengths.tex
% Generated by scripts/make_long_training_figures.py from frozen evidence; do not edit by hand.
\begin{tabular}{llrrrrrrrrr}
\toprule
& & \multicolumn{5}{c}{RULER} & \multicolumn{4}{c}{HELMET}\\
\cmidrule(lr){3-7}\cmidrule(lr){8-11}$\eta$ & $B$ & 4K & 8K & 16K & 32K & 64K & 8K & 16K & 32K & 64K\\
\midrule
\multicolumn{11}{l}{\textit{Qwen3.5}}\\
\multicolumn{2}{l}{Pretrained checkpoint} & 65.1 & 58.4 & 34.0 & 23.8 & 17.5 & 35.9 & 24.1 & 22.9 & 23.3\\
\addlinespace
3e-4 & 4M & 77.2 & 71.9 & 69.1 & 61.3 & 51.9 & 45.5 & 42.5 & 39.9 & 37.8\\
5e-4 & 4M & 78.8 & 74.3 & 70.1 & 61.6 & 52.6 & 45.7 & 41.7 & 39.1 & 38.4\\
1e-3 & 4M & 81.2 & 74.4 & 71.2 & 65.3 & 57.0 & 46.3 & 42.8 & 40.9 & 39.1\\
1.5e-3 & 4M & 79.2 & 75.0 & 72.2 & 65.6 & 56.0 & 47.7 & 45.5 & 42.1 & 40.0\\
3e-3 & 4M & 81.6 & 76.9 & 71.1 & 63.0 & 53.7 & 45.4 & 42.3 & 40.0 & 37.5\\
\addlinespace
1e-3 & 8M & 80.0 & 73.7 & 70.2 & 62.1 & 53.8 & 45.4 & 43.4 & 39.8 & 37.2\\
1.5e-3 & 8M & 80.4 & 74.4 & 70.5 & 64.7 & 54.7 & 46.6 & 44.6 & 41.0 & 38.3\\
3e-3 & 8M & 79.3 & 75.0 & 69.3 & 65.7 & 56.4 & 47.4 & 45.4 & 43.2 & 40.0\\
\midrule
\multicolumn{11}{l}{\textit{OLMo3}}\\
\multicolumn{2}{l}{Pretrained checkpoint} & 69.4 & 51.9 & 7.9 & 4.1 & 2.7 & 34.6 & 8.1 & 9.1 & 7.4\\
1e-3 & 4M & 75.3 & 63.2 & 52.4 & 43.3 & 36.1 & 42.1 & 37.1 & 34.7 & 29.6\\
\bottomrule
\end{tabular}

%% file: tables/context_categories.tex
\begin{table}[tp]
\caption{Qwen3.5 category scores at 64K; scores for the pretrained checkpoint are reference measurements. RULER categories are needle-in-a-haystack retrieval (Retr.), multi-hop tracing (Multi-hop), aggregation (Aggr.), and question answering (QA). HELMET categories are synthetic recall (Recall), retrieval-augmented generation (RAG), passage re-ranking (Rerank), many-shot in-context learning (ICL), and long-document QA (LongQA).}
\label{tab:context_categories}
\centering\footnotesize
\input{tables/generated/context_categories.tex}
\end{table}

%% file: tables/generated/context_categories.tex
% Generated by scripts/make_long_training_figures.py from frozen evidence; do not edit by hand.
\begin{tabular}{llrrrrrrrrr}
\toprule
& & \multicolumn{4}{c}{RULER} & \multicolumn{5}{c}{HELMET}\\
\cmidrule(lr){3-6}\cmidrule(lr){7-11}$\eta$ & $B$ & Retr. & Multi-hop & Aggr. & QA & Recall & RAG & Rerank & ICL & LongQA\\
\midrule
\multicolumn{2}{l}{Pretrained checkpoint} & 19.8 & 9.2 & 13.7 & 16.5 & 6.9 & 24.5 & 0.1 & 69.0 & 15.8\\
\midrule
3e-4 & 4M & 67.0 & 16.4 & 21.4 & 40.0 & 38.1 & 44.5 & 0.2 & 76.4 & 30.0\\
5e-4 & 4M & 68.7 & 17.0 & 21.6 & 37.0 & 40.9 & 44.8 & 0.4 & 73.8 & 32.1\\
1e-3 & 4M & 73.5 & 18.4 & 28.4 & 39.0 & 44.1 & 44.5 & 2.2 & 75.8 & 29.0\\
1.5e-3 & 4M & 71.5 & 18.2 & 27.3 & 42.0 & 41.2 & 45.7 & 2.4 & 78.0 & 32.9\\
3e-3 & 4M & 68.9 & 18.4 & 22.6 & 41.5 & 33.8 & 44.8 & 2.6 & 77.2 & 29.2\\
\midrule
1e-3 & 8M & 69.3 & 21.0 & 24.8 & 37.0 & 38.2 & 42.2 & 1.9 & 75.8 & 27.8\\
1.5e-3 & 8M & 70.0 & 20.8 & 25.6 & 40.0 & 41.8 & 45.2 & 1.3 & 74.4 & 28.6\\
3e-3 & 8M & 71.7 & 16.4 & 26.2 & 45.0 & 40.6 & 45.5 & 3.4 & 79.0 & 31.5\\
\bottomrule
\end{tabular}

%% file: tables/olmo3_long.tex
\begin{table}[tp]
\caption{OLMo3 results with the Qwen3.5 recipe reused without tuning. Base Main input limits change from 8K after pretraining to 64K after mid-training. Long-context scores for the pretrained checkpoint are reference measurements. Tables~\ref{tab:base_main_tasks} and~\ref{tab:context_lengths} give task-level Base Main results and long-context scores at each evaluation length, respectively.}
\label{tab:olmo3_long}
\centering\small
\input{tables/generated/olmo3_long.tex}
\end{table}

%% file: tables/generated/olmo3_long.tex
% Generated by scripts/make_long_training_figures.py from frozen evidence; do not edit by hand.
\begin{tabular}{lrrrrrrr}
\toprule
& \multicolumn{5}{c}{Base Main} & \multicolumn{2}{c}{64K}\\
\cmidrule(lr){2-6}\cmidrule(lr){7-8}Checkpoint & Score & Math & MC-STEM & MC-NonSTEM & GenQA & RULER & HELMET\\
\midrule
Pretrained checkpoint & 26.2 & 2.1 & 25.2 & 27.7 & 49.8 & 2.7 & 7.4\\
Mid-trained (1e-3/4M) & 40.5 & 16.4 & 43.8 & 49.2 & 52.4 & 36.1 & 29.6\\
\bottomrule
\end{tabular}

%% file: sections/16_appendix_literature_survey.tex
\section{Literature Survey Methodology}
\label{app:survey_methodology}

% Role: survey coverage and eligibility criteria.
The survey covers language-model architecture studies from November~2023 to October~2026, including multimodal models, with no minimum parameter count. Eligible papers described changes to a model's internal computation and reported training or evaluation experiments. Studies limited to data, optimization, prompting, or execution efficiency were excluded unless they included architectural changes.

% Role: retrieval and screening.
To identify eligible papers, we searched titles and abstracts across all arXiv categories, then used AI-assisted screening and targeted full-text checks.

% Role: group classification and affiliation sources.
Figure~\ref{fig:literature_survey} compares work from universities and independent researchers alone with work involving industry. We classified a paper as work involving industry if at least one author was affiliated with a company, government body, independent research institution, or non-university nonprofit, including dual affiliations. We obtained affiliations from formal publications or arXiv versions and also used verified information about authors' employment at the time of publication; funding alone did not determine the group.

% Role: date assignment and paper counts.
Each paper was counted once by publication date, using its earliest documented public release or its first arXiv submission when no earlier date was established. The figure spans \texttt{23Q4}--\texttt{26Q3}, with 2,919 papers in the paper-count plot. The three-month periods are labeled Q1 for February--April, Q2 for May--July, Q3 for August--October, and Q4 for November--January of the following year.

% Role: citation totals, matching, and display.
The citation plot contains 117,629 cumulative citations to 2,859 papers from the same periods. We retrieved citation counts from Semantic Scholar on October~31, 2026. We matched papers to Semantic Scholar records only when both the arXiv identifiers and normalized titles agreed; unmatched papers were excluded. In each plot, bars show group totals and the line shows the share of papers or citations for work involving industry. Citation totals are grouped by publication date and are unadjusted for paper count or age.